\documentclass[10pt,twocolumn,a4paper]{article}
\usepackage{preprint}
\usepackage{multirow}
\usepackage{placeins}
\usepackage{wrapfig}
\usepackage{longtable}
\usepackage{array}
\providecommand{\botrule}{\bottomrule}
\usepackage[pagebackref,breaklinks,colorlinks,allcolors=linkblue]{hyperref}

\input{generated/numbers}
\input{generated/family-numbers}
\input{generated/extra-numbers}
\input{generated/sens-numbers}

\title{\texorpdfstring{%
  \makebox[0pt][r]{\raisebox{-0.28\height}{\includegraphics[height=1.5em]{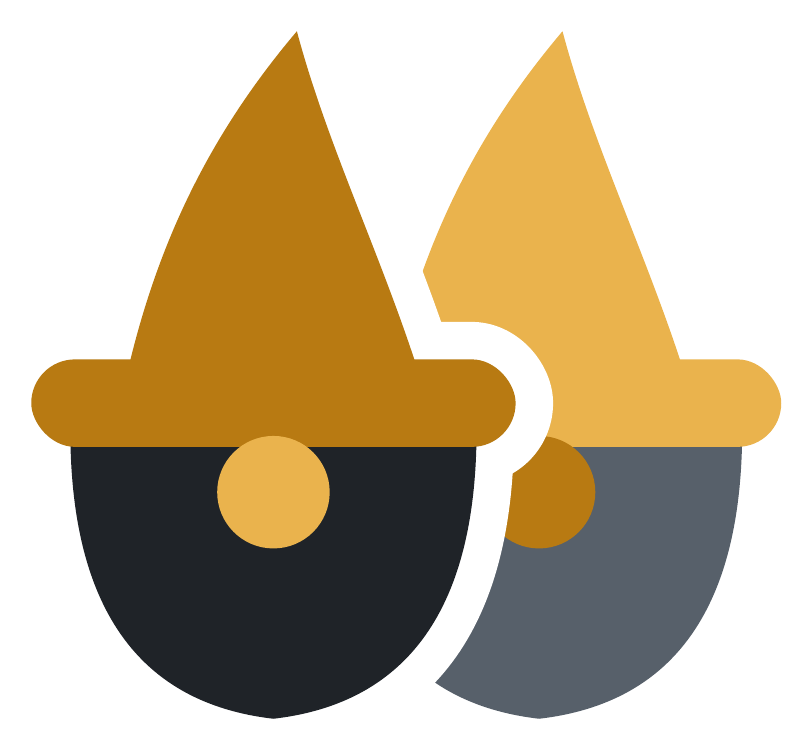}}\hspace{0.35em}}%
  Which Dwarf Is This?\\ Doppeldwarf-306 and Near-Identical Instance Retrieval}%
  {Which Dwarf Is This? Doppeldwarf-306 and Near-Identical Instance Retrieval}}

\author{Turhan Can Kargın\\
{\tt\small turhancan.kargin@gmail.com}\\[2pt]
{\small Project page: \url{https://turhancan97.github.io/doppeldwarf-306/}}}
\date{}

\hypersetup{pdfauthor={Turhan Can Kargın},
  pdftitle={Which Dwarf Is This? Doppeldwarf-306 and Near-Identical Instance Retrieval}}

\begin{document}
\maketitle

\begin{abstract}
Most established instance-retrieval benchmarks emphasise landmarks; small sculptures from visually similar series offer a complementary challenge.
We introduce Doppeldwarf-306: 1{,}691 openly licensed photographs of 306 bronze dwarf statues in Wroc{\l}aw, Poland, many made in themed series that share one style, with the photographer of each image recorded.
With no training, nearest-neighbour search over frozen DINOv2 features names the right statue first for \wv{dinov2_top_1}\% of leave-one-out queries, and the best of thirteen backbones from seven pretraining paradigms reaches \wv{fam_radio_top_1}\%.
Accuracy falls slowly as candidates are added: for DINOv2, by about 0.8 points each time their number doubles.
A natural design for a phone guide compares a photograph only with statues near the user.
In simulation this helps, but less than an equal number of randomly chosen statues does, so random subsampling overstates its value; a permutation test ties the penalty to the real layout ($p = \wv{dinov2_geo_permutation_mean_p_value}$).
Used as a weight instead of a filter, location raises DINOv2 from \wv{dinov2_soft_prior_none_top_1}\% to \wv{dinov2_soft_prior_gps25_top_1}\% for a simulated user 25\,m from the statue.
With the query's photographer withheld, learned local-feature matching raises DINOv2 accuracy from \wv{cv_cosine_top_1}\% to \wv{cv_loftr_top_1}\% and also improves DINOv3.
Comparing the two best candidates cuts DINOv2's false acceptance of unknown statues from \wv{dinov2_reject_cosine_false_acceptance_90}\% to \wv{dinov2_reject_margin_false_acceptance_90}\% at 90\% known acceptance.
\end{abstract}

\section{Introduction}
\label{sec:intro}

A visitor points a phone at a small bronze figure on a pavement and asks which one it is.
The city has hundreds of them, many made in themed series, and the visitor's photograph is the only evidence.
That question is instance-level retrieval: search a reference collection for the same physical object rather than the same kind of object~\cite{philbin2007oxford,radenovic2018revisiting,weyand2020gldv2}.
Frozen self-supervised features have made the first answer cheap.
Embeddings from models trained without labels transfer to retrieval well enough that nearest-neighbour search over them, with no fine-tuning, is a strong baseline rather than a straw man~\cite{radford2021clip,oquab2024dinov2}.

How far that baseline goes depends on what is being told apart, and the benchmarks that would answer it are dominated by landmarks and places~\cite{radenovic2018revisiting,weyand2020gldv2,berton2022benchmark}: objects that are large, distinctive by construction, and photographed from a distance.
Many real collections are the opposite.
Municipal sculpture, memorials and street furniture are frequently near-identical by design: cast from one model, commissioned as a themed series, or installed as variations on a single figure.
What distinguishes two members of such a family is small, local, and easily lost to a change of viewpoint or light.
Fine-grained recognition studies small differences, but between categories~\cite{krause2013cars,vanhorn2018inaturalist}; here every class is one physical statue or fixed installation, documented by a handful of photographs.

\begin{figure}[t]
  \centering
  \includegraphics[width=\linewidth]{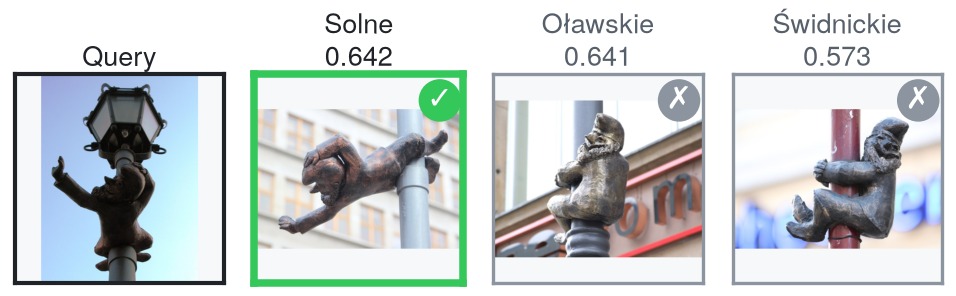}
  \caption{\textbf{Which dwarf is this?}
  A query photograph from one of three groups of pillar dwarves of one series, alike in style, scale and setting, installed on three streets, against the best-matching photograph of each group, with DINOv2 cosine similarity.
  The correct group wins by the third decimal place, through a different figure from the query's.
  The query is chosen by rule, not by eye: of this family's queries that DINOv2 answers correctly, it is the one with the smallest winning margin.}
  \label{fig:teaser}
\end{figure}

We study this regime on the bronze dwarves of Wroc{\l}aw, Poland: several hundred statues, many installed in families, photographed constantly by the public and partly released under free licences~\cite{szymanski2015orange,visitwroclaw2026dwarves}.
Three groups of pillar dwarves of one series stand on three different streets (\cref{fig:teaser}); two water-carrying dwarves share one plinth (\cref{fig:dataset}a), as does a sculpted blind and deaf pair.
We make three contributions.

\begin{itemize}
\item \textbf{Doppeldwarf-306, a corpus of near-identical instances, and how identification scales on it.}
1{,}691 openly licensed photographs of 306 statues, every image carrying its source, author and licence, and 294 statues placed on the map (\cref{sec:dataset}).
Recording the photographer of every image lets the evaluation withhold a query's own photographer, a stress test that the benchmarks in \cref{tab:positioning}, which do not record the photographer, cannot run.
Accuracy falls steadily as the candidate set grows from 2 statues to 306, approximately log-linearly for the strongest backbones (\cref{sec:narrow}).
\item \textbf{Location helps more as a weight than as a filter.}
In simulation, restricting candidates to the statues nearest the query statue scores below an equal-sized random restriction, for every backbone, and a permutation test shows that the penalty depends on who stands next to whom.
Weighting candidates by distance instead keeps every statue in play and lifts DINOv2 to \wv{dinov2_soft_prior_gps25_top_1}\% (\cref{sec:narrow}).
\item \textbf{A comparison of frozen representations.}
Thirteen frozen backbones across seven pretraining paradigms, compared on accuracy, scaling, geography, photographer dependence and rejection (\cref{sec:retrieve}).
Invariance-trained models (self-distillation and agglomerative distillation) lead on accuracy and rejection, while masked-prediction models (MAE, BEiT, I-JEPA) fail as frozen features under a fixed pooling rule.
\end{itemize}

\noindent
We organise the results by the stages of an identification system: retrieve, narrow, then verify and reject.
Lookalikes cause the most frequent confusions, but most errors are diffuse.

\section{Related Work}
\label{sec:related}

\paragraph{Instance retrieval and its benchmarks.}
Instance retrieval grew from quantised local features with spatial verification~\cite{sivic2003videogoogle,philbin2007oxford} through compact aggregation~\cite{jegou2010vlad,tolias2013asmk} to learned global descriptors~\cite{arandjelovic2016netvlad,noh2017delf,radenovic2019gem}, with query expansion as a standard refinement~\cite{chum2007expansion}.
Revisited Oxford and Paris~\cite{radenovic2018revisiting}, Google Landmarks v2~\cite{weyand2020gldv2} and place-recognition suites~\cite{berton2022benchmark,berton2022cosplace} emphasise buildings and scenes.
ILIAS~\cite{kordopatis2025ilias} instead covers 1{,}000 diverse objects at web scale and evaluates foundation models and local re-ranking.
Our contribution is a smaller outdoor-sculpture corpus with photographer provenance, look-alike series and controlled random-versus-geographic candidate pools.
Where geo-localization learns representations using distance~\cite{han2026swclip,cui2026geometric}, we ask how a known position should enter a frozen instance ranking.

\paragraph{Frozen features and local matching.}
DINOv2~\cite{oquab2024dinov2}, CLIP~\cite{radford2021clip} and SigLIP~2~\cite{tschannen2025siglip2} support transfer without dataset-specific training.
Reports of weaker nearest-neighbour performance from masked-prediction patch features~\cite{vanyan2024local} motivate our global comparison under a fixed pooling rule.
Verification is an established retrieval refinement~\cite{philbin2007oxford,kordopatis2025ilias}; we test SIFT~\cite{lowe2004sift}, DISK+LightGlue~\cite{tyszkiewicz2020disk,lindenberger2023lightglue} and LoFTR~\cite{sun2021loftr} on bronze sculpture with photographer withholding and held-out blend weights, and compare geometry alone as a first stage.
Multi-vector descriptors address small objects or local detail~\cite{green2025needle,bezerra2026regionaware,choi2026patchwise}; we test a simple training-free grid variant.

\paragraph{Lookalikes and heritage.}
Fine-grained recognition separates similar categories~\cite{nilsback2008flowers,krause2013cars,vanhorn2018inaturalist,wei2022finegrained}; here each class is one physical statue or installation.
Heritage collections pose instance-level recognition directly, over museum holdings~\cite{mensink2014rijksmuseum}, painting databases~\cite{seguin2016paintings,shen2019artpatterns} and art collections~\cite{ufer2020artretrieval,castellano2021artsurvey}.
Closest to ours, The Met~\cite{ypsilantis2022met} tests artwork instances using studio references and visitor queries.
Doppelgangers~\cite{cai2023doppelgangers}, like our corpus, draws outdoor photographs from Wikimedia Commons, but targets pairwise disambiguation of similar structures for reconstruction.
We rank statue identities and measure contributor overlap and spatial layout effects on retrieval.
Public-image coverage and duplicates can bias evaluation~\cite{torralba2011bias,barz2020duplicates}; recording each image's photographer enables a direct contributor-overlap stress test (\cref{sec:protocol}).

\section{Doppeldwarf-306: A Corpus of Near-Identical Instances}
\label{sec:dataset}

\begin{figure*}[t]
  \centering
  \includegraphics[width=\linewidth]{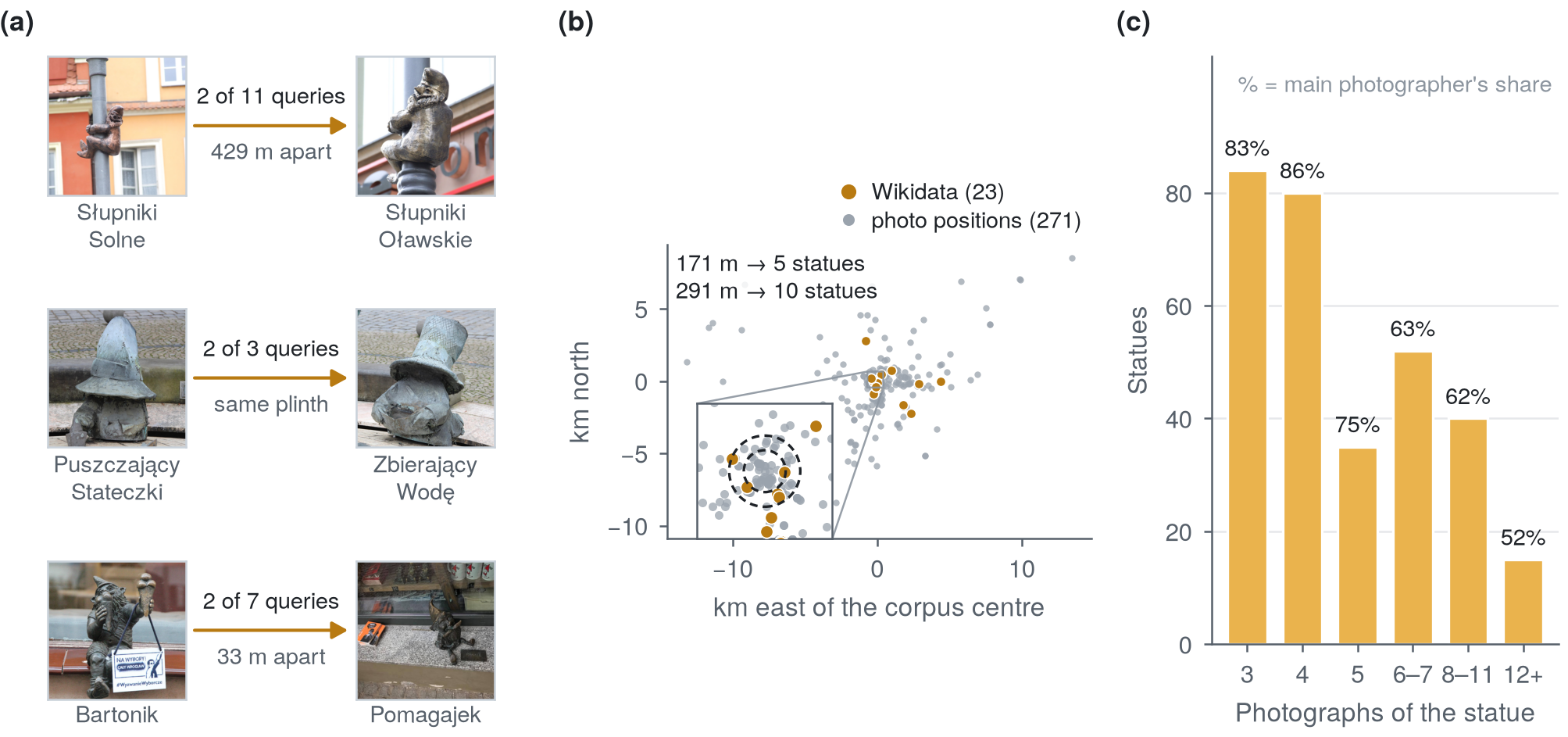}
  \caption{\textbf{The corpus and the three difficulties it presents.}
  (a)~DINOv2's three most frequent confusions between different sculptures, ranked by misidentifications and then by how often the pair competes (a group class against its own member is skipped, and each statue appears once); each arrow gives how many of the left statue's queries are returned as the right one.
  S{\l}upniki Solne and S{\l}upniki O{\l}awskie stand 429\,m apart in a series of three pillar-dwarf groups; the second pair shares one plinth, and the third stands 33\,m apart.
  (b)~The 294 of 306 statues that carry coordinates, in kilometres from the corpus centre: 23 from Wikidata, the rest the median of the camera positions in a statue's own photographs; inset, the densest 300\,m neighbourhood at the two radii a location-aware guide would use.
  (c)~Statues by number of photographs, each bar labelled with the mean share of a class's photographs taken by its most prolific photographer: two-thirds of classes have five photographs or fewer, and the smallest draw 83\% of them from one person.}
  \label{fig:dataset}
\end{figure*}

\paragraph{Sources and review.}
Statues were found through the Wikimedia Commons category tree, which holds 481 per-statue categories; Wikidata~\cite{vrandecic2014wikidata} has an item for only 44 of them.
Every candidate mapping from a category to a statue carries a human review decision: regular titles were approved by a naming rule and the rest inspected individually, giving 469 approved and 13 rejected (details in the appendix).
A category that shows one fixed installation of several figures is one class; 46 of the final 306 classes are such installations.
Images were admitted only under licences that permit redistribution (public domain, CC0, CC~BY, CC~BY-SA), and every image carries its source, author and licence.
These licences cover the photographs, not the sculptures, which are contemporary works photographed under Poland's freedom of panorama; photographs are removed on request.
Stored copies have a short side of at least 400 and a long side of at most 2{,}000 pixels.

\paragraph{Leakage control.}
36 Commons files were byte-identical under \emph{different} statue labels, almost always a group category and one of its members holding the same photograph.
Such a file would put a query's answer in the reference set under a second name, so every copy is dropped.
Requiring at least three photographs per statue, so that leave-one-out is possible, leaves 306 of the 469 approved mappings: 1{,}691 photographs by 122 photographers, with a median of four per statue and a maximum of 31 (\cref{fig:dataset}c).
The corpus is therefore bounded by documentation: of the several hundred statues in the city's public space~\cite{visitwroclaw2026dwarves}, it holds those documented on Commons with at least three usable, freely licensed photographs.

\paragraph{Locations.}
Wikidata places 23 statues.
For the rest we take the median of the camera positions recorded in each statue's own photographs, on the reasoning that a 30\,cm bronze is photographed from a few metres away.
On the 21 statues that carry both, the two disagree by a median of 9\,m, well below the 52\,m median radius of the smallest geographic pool (two statues) and the 171\,m of a five-statue pool.
294 of the 306 statues are placed; the twelve without a usable coordinate are excluded from the geographic comparison only.

\paragraph{Photographers.}
Contribution is concentrated: two of the 122 photographers supplied 67.4\% of the images, and 125 statues were photographed by a single person.
We keep this as a property of the corpus and measure its cost (\cref{sec:protocol}).

\begin{table}[t]
  \centering
  \footnotesize
  \setlength{\tabcolsep}{2.5pt}
  \begin{tabular}{@{}lllr@{}}
    \toprule
    Benchmark & Objects & Level & Scale \\
    \midrule
    $\mathcal{R}$Oxf/Par~\cite{radenovic2018revisiting} & Landmarks & Inst. & 5k/6k imgs \\
    GLDv2~\cite{weyand2020gldv2} & Landmarks & Inst. & 200k inst. \\
    The Met~\cite{ypsilantis2022met} & Artworks & Inst. & 224k cl. \\
    Flowers~\cite{nilsback2008flowers} & Species & Categ. & 103 cl. \\
    iNat~\cite{vanhorn2018inaturalist} & Species & Categ. & $>$5k cl. \\
    \midrule
    Ours & Sculpture & Inst. & 306 inst. \\
    \bottomrule
  \end{tabular}
  \caption{\textbf{Where the corpus sits.} Scales as reported in each work's abstract.
  None of the others records who took each photograph; ours does, and uses it.
  It does not compete on size; it is built around look-alike series.}
  \label{tab:positioning}
\end{table}

\section{Protocol}
\label{sec:protocol}

\paragraph{Retrieval.}
We use four frozen backbones: DINOv2 ViT-B/14~\cite{oquab2024dinov2} (CLS token, 768-d), CLIP ViT-B/32~\cite{radford2021clip} (projected image embedding, 512-d), DINOv3 ViT-B/16~\cite{simeoni2025dinov3} (CLS token, 768-d), and SigLIP~2 ViT-B/16~\cite{tschannen2025siglip2} (pooled image embedding, 768-d), a current language-aligned model of the same size as the two self-supervised ones.
No backbone parameter is fitted to the corpus; each is pinned to an immutable revision and uses the image processor it was released with.
DINOv2 is the reference throughout: the re-ranking analyses use its candidate lists, the three matchers are also re-run on DINOv3's (\cref{sec:verify}), and DINOv3 is reported wherever it can be compared.
To compare released checkpoints across pretraining paradigms, we add nine frozen backbones, for thirteen across seven pretraining paradigms (roster in the appendix): joint-embedding self-distillation (DINO~\cite{caron2021dino}, DINOv2, DINOv3), a joint-embedding predictive architecture (I-JEPA~\cite{assran2023ijepa}), masked image modelling (MAE~\cite{he2022mae}, BEiT~\cite{bao2022beit}, EVA-02~\cite{fang2024eva02}), vision--language alignment (CLIP B/32 and B/16, SigLIP~2), agglomerative distillation (C-RADIOv3~\cite{ranzinger2024amradio,heinrich2025radiov25}), autoregressive pretraining (AIMv2~\cite{fini2025aimv2}), and a supervised ViT trained on ImageNet-21k labels~\cite{steiner2022augreg} as a control.
We use ViT-B wherever it is released; I-JEPA (H/14) and AIMv2 (L/14) are not, C-RADIOv3 reads 432-pixel inputs, AIMv2 also predicts captions, and EVA-02 reconstructs CLIP features.
Each is pooled by one rule fixed in advance: its released embedding where it has one, otherwise the mean of its patch tokens. Masked-prediction features are known to serve poorly when frozen and pooled~\cite{he2022mae,balestriero2024reconstruction,vanyan2024local}, so their results describe those models under this rule.
With one to three models per paradigm, and training data differing between them, differences between paradigms are descriptive; intervals are per model.
Vectors are L2-normalised and searched exhaustively by cosine similarity.
Candidates are collapsed from photographs to statues, each statue ranked by its best-matching photograph, because that is the list an identification tool would show.
A query is never compared against a byte-identical copy of itself.

\paragraph{Leave-one-out.}
Each of the 1{,}691 photographs is queried against the other 1{,}690, and an answer is correct when the query's own statue is ranked first among the 306.
Every statue has at least three photographs, so at least two of the correct statue always remain.
This is the standard protocol and an optimistic one: the remaining photographs often come from the same photographer, which the next test removes.

\paragraph{Unseen-photographer stress test.}
A visitor is a photographer the reference set has never seen.
We simulate this by withholding, for each query, every photograph by the same photographer.
The 125 statues photographed by a single person then have no correct reference left, so this arm scores the remaining 1{,}157 queries, and every drop we report from it compares arms on those same 1{,}157.
Withholding a photographer also withholds references, which lowers accuracy for reasons unrelated to who held the camera.
A size-matched control therefore removes references at random over five seeds, matching both the total number kept and the number of correct-statue references kept.
The gap between the disjoint arm and the control is what we attribute to the photographer.
This is a controlled stress test of contributor overlap; it does not measure the gap to casual street photographs, which no query here represents (\cref{sec:discussion}).

\paragraph{Pool sampling.}
Each query is scored against its own statue plus $N-1$ others drawn at random, over five seeds at each of eleven sizes from 2 to 306.
The geographic variant takes the query statue's $N-1$ nearest placed neighbours instead.
That pool is deterministic, so it is compared against the five-seed mean of a random pool of the same size over the same 294 placed statues. Both are simulations centred on the true statue, and the random pool, which always contains it, is a baseline that a deployed guide could not use.

\paragraph{Re-ranking.}
For each of the first stage's top ten statues we match local features between the query and that statue's best photograph, fit a RANSAC homography~\cite{fischler1981ransac} with a 5\,px tolerance, and blend the inlier count $n$ into the cosine score $s$ as $s + w \min(n, 30)/30$; the cap of 30 was fixed for SIFT before any learned matcher was tried, and the appendix varies it.
Ties keep the cosine order, so $w=0$ reproduces the first stage exactly. A homography only approximates the view change of a 3D statue; the inlier count serves only as a ranking signal.
SIFT~\cite{lowe2004sift} uses up to 800 keypoints and DISK~\cite{tyszkiewicz2020disk} matched by LightGlue~\cite{lindenberger2023lightglue} up to 1{,}024; the detector-free LoFTR~\cite{sun2021loftr} matches images at a long side of 840\,px and keeps its 1{,}024 most confident matches above its released threshold. As a check on the shortlist, SIFT and DISK are also run on the top 50 (appendix). The appendix lists the fixed weight grid.
Choosing $w$ on the queries it is then scored on would flatter the result, so we choose it by five-fold cross-validation over \emph{statues}.
The statues are split deterministically into five groups, $w$ is chosen on four (ties go to the smaller weight), and it is applied to the fifth.
Every query is scored exactly once, by a weight chosen without any photograph of its statue.
Held-out statues remain in the reference gallery; only the selection of $w$ is separated from them.
The protocol and grid were fixed while exploring these queries, so this is a validation check on explored data.

\paragraph{Location as a weight.}
The geographic variant uses location as a hard filter. As a soft alternative, every placed statue is ranked by its cosine score plus $\lambda\,e^{-d/200\,\mathrm{m}}$, where $d$ is its distance from the user. The user is placed 10, 25, 50 or 100\,m from the query statue in a random direction (five draws), because at zero error a strong prior simply names the statue underfoot. $\lambda$ is chosen by the same statue-level cross-validation as $w$, and the statue nearest the user is reported as a location-only baseline. Neither the 200\,m scale nor the exponential shape was tuned, and the appendix varies both.

\paragraph{Other first-stage changes.}
We also test a linear probe and class prototypes fitted on each leave-one-out fold, fused DINOv2 and CLIP scores, query expansion, DINOv2 with registers~\cite{darcet2024registers} and DINOv2 ViT-L/14 (3.5$\times$ the parameters), and ranking every reference by inlier count alone. Recall at ten bounds any top-ten re-ranker; inlier ties are reported as a best-to-worst range. Results: \cref{sec:negative} and the appendix.

\paragraph{Rejection.}
A query is \emph{unknown} when every photograph of its statue is removed from the gallery, giving 1{,}691 unknown queries beside the 1{,}691 known ones; a known query is handled correctly only if it is accepted with its statue ranked first.
Two training-free scores are thresholded: the top-1 cosine, and the margin between the best and the second-best statue.
Each threshold is calibrated leave-one-class-out, as a quantile of the other statues' known scores, so no query helps set the threshold that judges it.
We report the AUROC between known and unknown scores and FA@90, the share of unknown queries accepted when 90\% of known ones are.

\paragraph{Testing the explanations.}
Two tests probe the account given for the geographic result. A permutation test shuffles which statue stands at which of the 294 positions (200 permutations), which keeps the random arm and the set of positions and breaks only who stands next to whom. Its $p$-value is the share of shuffled layouts at least as harmful as the real one. A sensitivity test moves every coordinate 9, 50 or 150\,m in a random direction. Lookalike families are defined without reference to any one model's errors: two statues are linked when each is among the other's three nearest by mean embedding under at least two backbones, families are the connected groups, and a backbone's errors are judged only against families drawn by the \emph{other} backbones. A first definition, mutual nearest neighbours, could only form pairs; it was replaced by this one, which was fixed before its own result was seen.

\paragraph{Uncertainty.}
Photographs of one statue share a sculpture, often a photographer and a visit, so they are not independent.
Our intervals are 95\% percentile intervals from 10{,}000 bootstrap resamples of whole statues, with matching draws across the methods or conditions being compared.
Point estimates stay per photograph.
For held-out re-ranking the intervals condition on the chosen folds.
Paired comparisons also report discordant counts with an exact McNemar test~\cite{mcnemar1947,dietterich1998}, which treats photographs as independent and so complements the statue bootstrap.

\section{Results}
\label{sec:results}

\subsection{Retrieve: comparing frozen representations}
\label{sec:retrieve}

\IfFileExists{figures/fig-family.png}{%
\begin{figure*}[t]
  \centering
  \includegraphics[width=\linewidth]{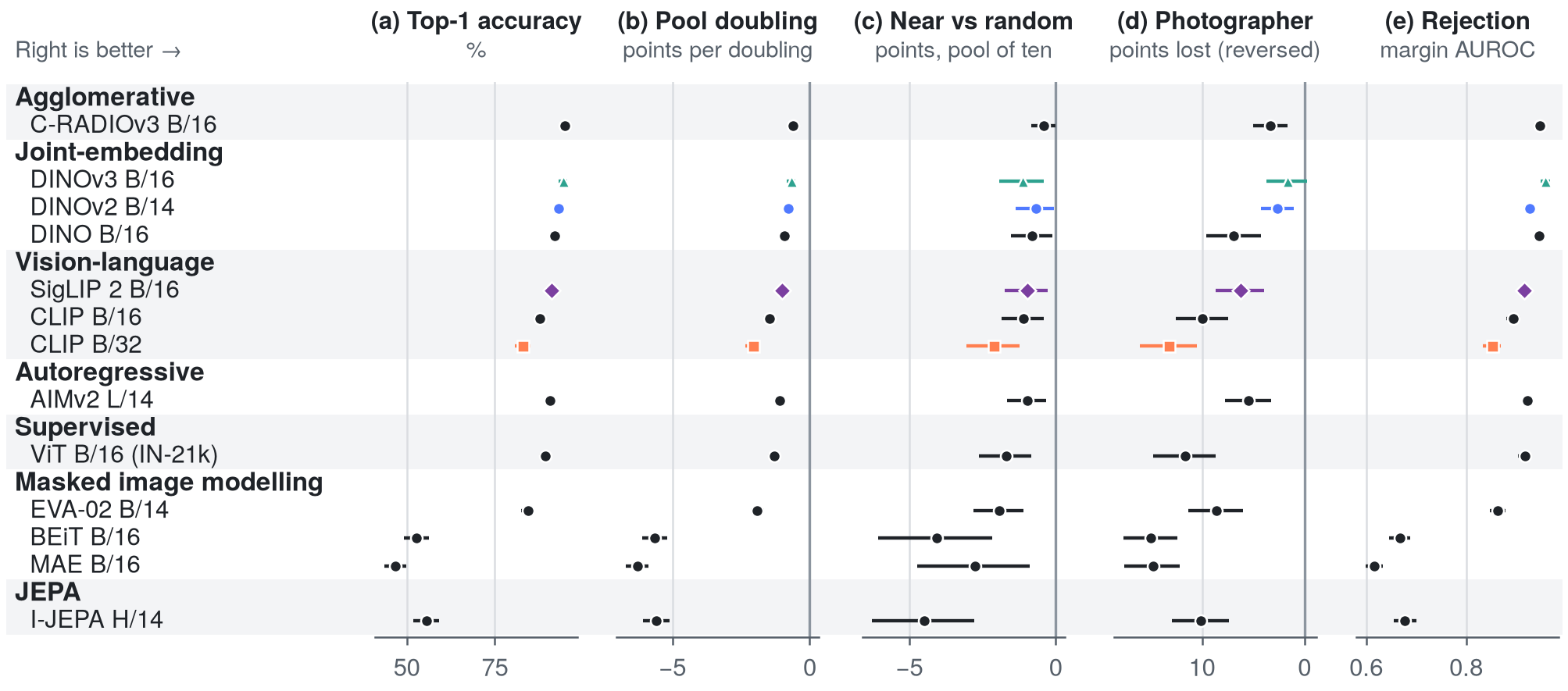}
  \caption{\textbf{Thirteen frozen backbones by pretraining paradigm, best top-1 first; right is better in every panel.}
  (a)~Top-1 over all 306 statues. (b)~Top-1 change per doubling of the candidate pool. (c)~Top-1 of a geographic pool of ten minus an equal-sized random pool. (d)~Top-1 lost to the photographer beyond the size-matched control, on a reversed axis. (e)~Rejection AUROC of the margin score.
  Bars are 95\% intervals from resampling whole statues; models are ViT-B unless named, and the four backbones used elsewhere in the paper keep their colours.}
  \label{fig:family}
\end{figure*}}{}

\begin{table}[t]
  \centering
  \footnotesize
  \setlength{\tabcolsep}{3.5pt}
  \begin{tabular}{@{}lrrrrr@{}}
    \toprule
     & & Unseen & Per & \multicolumn{2}{c}{FA@90} \\
    \cmidrule(l){5-6}
    Backbone & Top-1 & ph. & doubling & cosine & margin \\
    \midrule
    DINOv2 & \wv{fam_dinov2_top_1} & \wv{fam_dinov2_unseen_top_1} & \wv{fam_dinov2_points_per_doubling} & \wv{fam_dinov2_cosine_false_acceptance_90} & \wv{fam_dinov2_reject_margin_false_acceptance_90} \\
    CLIP & \wv{fam_clip_top_1} & \wv{fam_clip_unseen_top_1} & \wv{fam_clip_points_per_doubling} & \wv{fam_clip_cosine_false_acceptance_90} & \wv{fam_clip_reject_margin_false_acceptance_90} \\
    DINOv3 & \wv{fam_dinov3_top_1} & \wv{fam_dinov3_unseen_top_1} & \wv{fam_dinov3_points_per_doubling} & \wv{fam_dinov3_cosine_false_acceptance_90} & \wv{fam_dinov3_reject_margin_false_acceptance_90} \\
    SigLIP~2 & \wv{fam_siglip2_top_1} & \wv{fam_siglip2_unseen_top_1} & \wv{fam_siglip2_points_per_doubling} & \wv{fam_siglip2_cosine_false_acceptance_90} & \wv{fam_siglip2_reject_margin_false_acceptance_90} \\
    \bottomrule
  \end{tabular}
  \caption{\textbf{The four main backbones at a glance.} Top-1 (\%) on all 1{,}691 queries and, photographer withheld, on the 1{,}157 answerable ones; points per doubling of the pool; unknown statues accepted (\%) at 90\% known acceptance.}
  \label{tab:glance}
\end{table}

Against all 306 statues, DINOv2 places the correct statue first for 93.1\% of queries (95\% statue interval \wvci{dinov2_top_1}), CLIP for 82.9\% (\wvci{clip_top_1}) and DINOv3 for 94.4\% (\wvci{dinov3_top_1}); \cref{tab:glance} collects the headline figures.
CLIP's deficit is not inherent to language alignment: SigLIP~2, a newer language-aligned model of the same size that also uses self-supervised losses, reaches \wv{siglip2_top_1}\% (\wvci{siglip2_top_1}).
\Cref{fig:family} widens the comparison to thirteen frozen backbones in seven pretraining paradigms, and three groups emerge.
Models trained for invariant embeddings lead on accuracy: the self-distillation line, from DINO (\wv{fam_dino_top_1}\%) to DINOv3, and C-RADIOv3 (\wv{fam_radio_top_1}\%, \wvci{fam_radio_top_1}), which distils DINOv2 among its teachers.
Language-aligned, autoregressive and supervised models follow, from CLIP B/32 to SigLIP~2, AIMv2 (\wv{fam_aimv2_top_1}\%) and the supervised ViT (\wv{fam_vit-supervised_top_1}\%); within language alignment, a smaller patch size closes about half the gap (CLIP B/16: \wv{fam_clip-b16_top_1}\%).
Masked-prediction models fail as frozen features under our pooling rule: MAE reaches \wv{fam_mae_top_1}\%, BEiT \wv{fam_beit_top_1}\% and I-JEPA, which predicts latent features rather than pixels, \wv{fam_ijepa_top_1}\% despite its H/14 size, consistent with reports that such features need fine-tuning or feature surgery for nearest-neighbour use~\cite{he2022mae,vanyan2024local}.
Masking alone does not decide it: EVA-02 is masked-image-modelled too, but regresses the features of a contrastively trained CLIP model, and reaches \wv{fam_eva02-mim_top_1}\%.
The other panels broadly follow the same order: the most accurate models also decay slowest and reject best, though not strictly: DINO B/16 is no less photographer-dependent than SigLIP~2 or AIMv2, and DINOv3's geographic penalty is larger than DINOv2's.
The appendix lists top-5, MRR (which equals mAP here) and unseen-photographer recall at ten for all thirteen.

\begin{figure}[t]
  \centering
  \includegraphics[width=\linewidth]{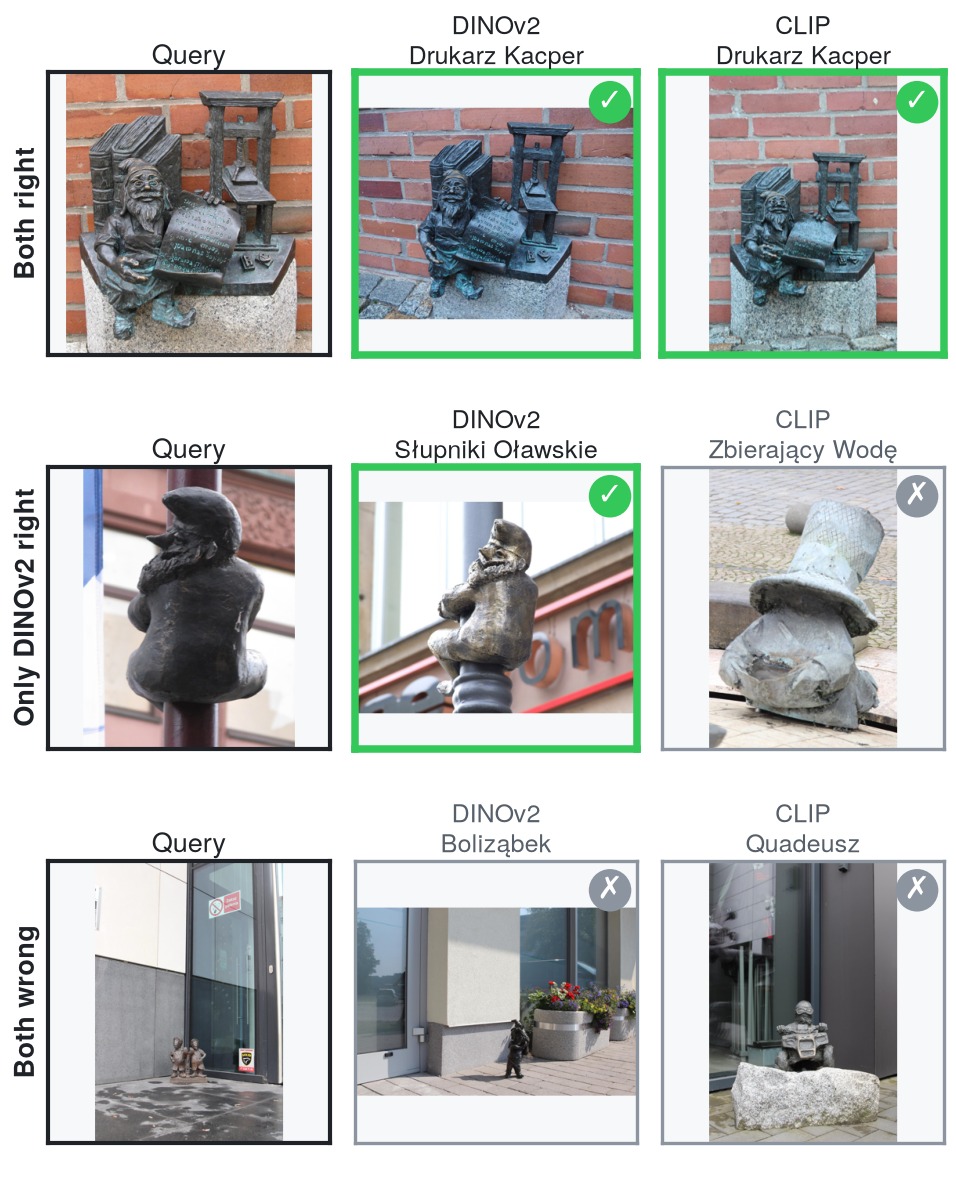}
  \caption{\textbf{Retrieval examples.} Each row is a query (left) and the statue each backbone ranks first; a green frame and check mark is correct, a grey frame and cross wrong.
  The rows are the first queries, in image-ID order, where both backbones are right, where only DINOv2 is right, and where both are wrong.
  CLIP's error in the middle row is a water-installation dwarf; in the bottom row the statue occupies a few dozen pixels of a facade.
  The top five for each, and credits for every photograph the paper shows, are in the appendix.}
  \label{fig:examples}
\end{figure}

DINOv2 misidentifies 116 queries and CLIP 289.
The most frequent individual confusions are between lookalike statues.
The three S{\l}upniki, groups of pillar dwarves of one series installed on three streets, are the top confusion for DINOv2 and CLIP, in both directions.
Among different sculptures they are followed by two water-carrying dwarves on one plinth and a pair standing 33\,m apart (\cref{fig:dataset}a).
DINOv3's top confusion is a sculpted blind and deaf pair on one plinth, and SigLIP~2's a group statue and one of its own members, a labelling overlap discussed below.
Lookalikes do not, however, account for most errors.
Defining families without naming them by hand (\cref{sec:protocol}) puts \wv{dinov2_families_statues} statues, \wv{dinov2_families_statue_share}\% of the corpus, in the families against which DINOv2 is judged (those drawn by the other three backbones).
Only \wv{dinov2_families_error_share}\% of DINOv2's errors are confusions within a family (\wvci{dinov2_families_error_share}; CLIP \wv{clip_families_error_share}\%, DINOv3 \wv{dinov3_families_error_share}\%, SigLIP~2 \wv{siglip2_families_error_share}\%), and statues in families err no more often than the rest (\wv{dinov2_families_inside_error_rate}\% against \wv{dinov2_families_outside_error_rate}\% for DINOv2; appendix).
Most errors are diffuse instead.
Some statues occupy a small part of a frame dominated by background (\cref{fig:examples}, bottom), and two group classes overlap their own members by construction (\eg \emph{Grajek i Meloman} alongside \emph{Grajek} and \emph{Meloman}).
Counting a group and its members as one changes only \wv{dinov2_groups_changed_queries} DINOv2 queries (\wv{dinov2_groups_merged_gain} points, \wvci{dinov2_groups_merged_gain}), so the overlap barely inflates the error count.

\paragraph{Unseen photographer.}
On the 1{,}157 answerable queries, withholding the query's photographer lowers DINOv2 from 94.2\% to 81.8\%.
The size-matched control alone reaches 84.5\%, so most of the drop comes from having fewer references.
The part attributable to the photographer is 2.6 points for DINOv2 (\wvci{dinov2_photographer_attributable}) and 13.2 for CLIP (\wvci{clip_photographer_attributable}).
For DINOv3 it is 1.7 points (\wvci{dinov3_photographer_attributable}), an interval that includes zero, and for SigLIP~2, \wv{siglip2_photographer_attributable} points (\wvci{siglip2_photographer_attributable}): language alignment carries part of CLIP's dependence on the photographer, but far less than CLIP suggests.
Scarcity is visible directly: with one reference of the right statue left, DINOv2 scores \wv{r2_dinov2_strata_1_top_1}\% (\wvci{r2_dinov2_strata_1_top_1}), with six or more \wv{r2_dinov2_strata_6plus_top_1}\% (\wvci{r2_dinov2_strata_6plus_top_1}).

\subsection{Narrow: location helps more as a weight than as a filter}
\label{sec:narrow}

\begin{figure}[t]
  \centering
  \includegraphics[width=\linewidth]{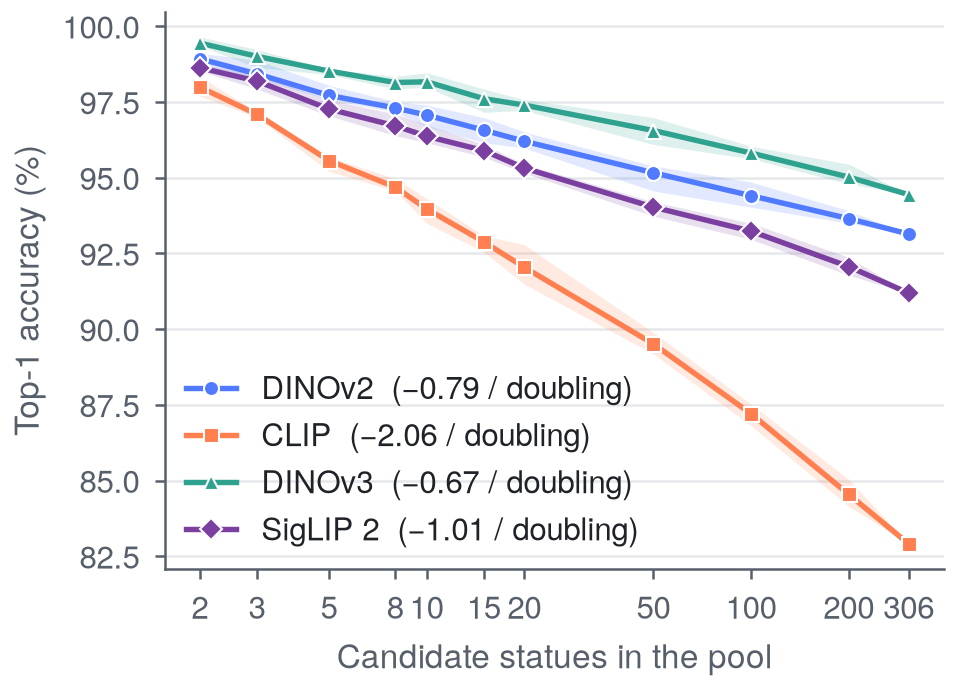}
  \caption{\textbf{Accuracy falls as candidates are added.}
  Each query against its own statue plus $N-1$ random others, five seeds per size; the band is the spread across seeds.
  The four backbones used throughout the paper; \cref{fig:family}b gives the decay for all thirteen, and the appendix their full curves.}
  \label{fig:pool}
\end{figure}

\paragraph{The scaling curve.}
Fitted over the eleven pool sizes, the change in top-1 per doubling of the pool is \wv{dinov2_points_per_doubling} points for DINOv2 (\wvci{dinov2_points_per_doubling}), \wv{clip_points_per_doubling} for CLIP (\wvci{clip_points_per_doubling}), \wv{dinov3_points_per_doubling} for DINOv3 (\wvci{dinov3_points_per_doubling}) and \wv{siglip2_points_per_doubling} for SigLIP~2 (\wvci{siglip2_points_per_doubling}) (\cref{fig:pool}).
DINOv2's decay is approximately log-linear: a loss of 0.8 points per doubling both between 2 and 20 statues and between 50 and 306.
CLIP's accelerates, from 1.8 to 2.5, so a single slope understates its loss at the large end.

\begin{figure*}[t]
  \centering
  \includegraphics[width=\linewidth]{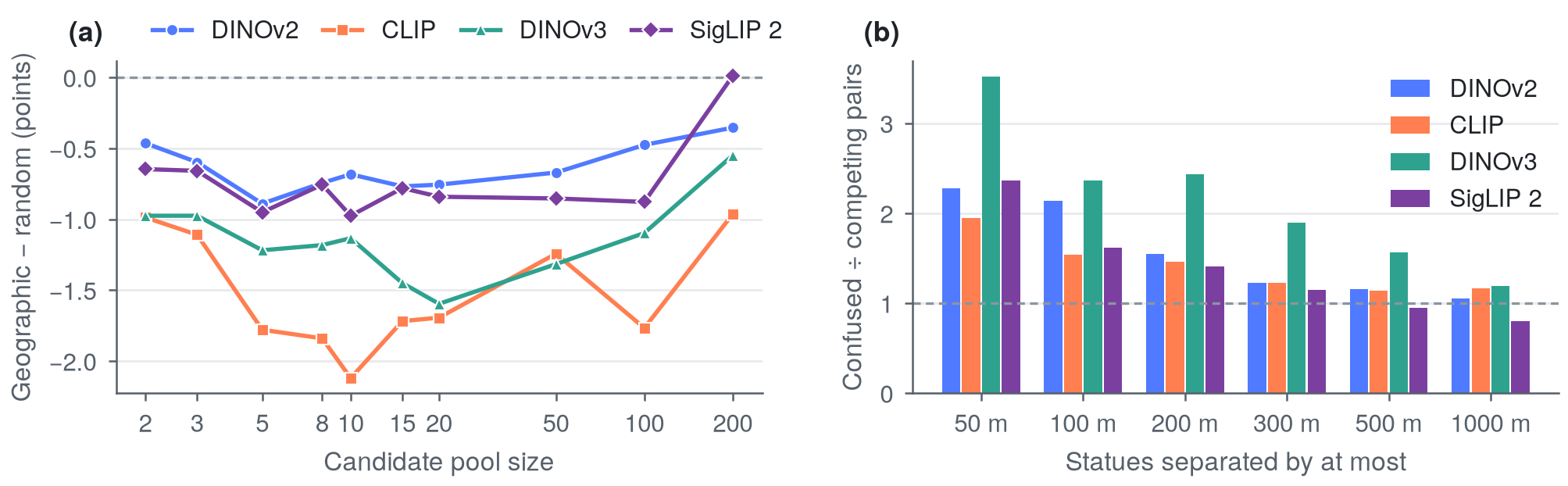}
  \caption{\textbf{Narrowing by location underperforms narrowing at random.}
  (a)~Top-1 of a pool of the query statue's geographic neighbours minus an equal-sized random pool, over the 294 placed statues.
  (b)~Pairs the retrieval confuses divided by pairs that merely compete, by separation distance; the dashed line is no enrichment.
  The four backbones used throughout the paper; \cref{fig:family}c gives the penalty at a pool of ten for all thirteen, and the appendix their full curves.}
  \label{fig:geo}
\end{figure*}

\paragraph{Location narrowing.}
The obvious design for a phone-based guide is to rank only the statues near the user.
A geographic pool of the query statue's nearest neighbours scores below an equal-sized random pool at every size from 2 to 100 for all four backbones, and at 200 for all but SigLIP~2, whose difference there is zero (\cref{fig:geo}a).
The statue intervals exclude zero at every size from 2 to 200 for CLIP and DINOv3, at eight of ten for DINOv2 (the exceptions are pools of 2 and 100), and at every size up to 100 for SigLIP~2.
At a pool of ten the penalty holds for all thirteen backbones of \cref{fig:family}c, every interval below zero except C-RADIOv3's, whose upper end touches zero.
The largest differences fall at the pool sizes such a guide would use: $-0.9$ points for DINOv2 at five statues (\wvci{dinov2_geo_minus_random_pool_5}; a 171\,m radius), $-2.1$ for CLIP at ten (\wvci{clip_geo_minus_random_pool_10}; 291\,m) and $-1.6$ for DINOv3 at twenty (\wvci{dinov3_geo_minus_random_pool_20}; 466\,m).
Geographic narrowing still improves on the full pool; it improves less than random narrowing does.

Confused pairs are enriched at short distances.
Among pairs of statues that compete for the same query, pairs the retrieval actually confuses are 2.1 times as likely to stand within 100\,m of each other (11.9\% against 5.5\% for DINOv2), and 2.3 times as likely within 50\,m (\cref{fig:geo}b).
The enrichment is short-range: by a kilometre the ratio is 1.06, proximity alone barely separates confused from competing pairs (AUROC 0.518), and 88\% of confused pairs stand more than 100\,m apart.
A neighbourhood pool retains nearby confusable statues while discarding distant distractors; random sampling therefore overstates the gain from geographic narrowing in this corpus.

The penalty depends on the association between statue identity and position. Shuffling which statue stands where, while keeping the positions, removes the penalty: averaged over pools of 5, 10 and 20, the real layout changes DINOv2's top-1 by \wv{dinov2_geo_permutation_mean_observed} points against \wv{dinov2_geo_permutation_mean_null} for shuffled layouts (95\% of shuffles \wvci{dinov2_geo_permutation_mean_null}).
No shuffled layout was as harmful as the real one for any of the four backbones ($p = \wv{dinov2_geo_permutation_mean_p_value}$, the smallest value \wv{dinov2_geo_permutation_permutations} permutations allow).
This does not isolate sculptural similarity from shared backgrounds or contributor patterns.
The five-statue penalty also survives coordinate error: moving every statue 150\,m in a random direction leaves it at \wv{dinov2_geo_minus_random_pool_5_noise_150} points (\wvci{dinov2_geo_minus_random_pool_5_noise_150}; 9 and 50\,m: appendix).

Adding $\lambda e^{-d/200\,\mathrm{m}}$ to each statue's score, with $\lambda$ chosen on held-out statues and the user 25\,m from the statue, lifts DINOv2 over the 294 placed statues from \wv{dinov2_soft_prior_none_top_1}\% to \wv{dinov2_soft_prior_gps25_top_1}\% (\wvci{dinov2_soft_prior_gps25_gain} points), and CLIP from \wv{clip_soft_prior_none_top_1}\% to \wv{clip_soft_prior_gps25_top_1}\%.
An oracle-centred filter of ten reaches 96.2\% and an equal random pool 96.9\%.
Location alone names the right statue \wv{dinov2_soft_prior_gps25_location_only_top_1}\% of the time, so the two cues are complementary.
At 25\,m the gain barely depends on the kernel's shape or scale (\wv{sens_prior_gps25_min_top_1}\% to \wv{sens_prior_gps25_max_top_1}\%); at 100\,m a wider kernel holds more of it (appendix).
The soft prior retains all candidates and uses distance to distinguish candidates within a neighbourhood; the hard filter uses only membership. This comparison does not isolate retaining candidates from adding that distance information.

\subsection{Verify and reject}
\label{sec:verify}
\label{sec:negative}
\label{sec:reject}

\begin{table}[t]
  \centering
  \footnotesize
  \setlength{\tabcolsep}{3pt}
  \caption{\textbf{Held-out verification.} DINOv2, 1{,}157 unseen-photographer queries over 181 statues. Top-1 is in percent; gains are points with paired 95\% statue intervals. Weights are selected on the other four statue folds.}
  \label{tab:verification}
  \begin{tabular}{@{}lrr@{}}
    \toprule
    Method & Top-1 & Gain [95\% CI] \\
    \midrule
    Cosine & \wv{cv_cosine_top_1} & --- \\
    + SIFT & \wv{cv_sift_top_1} & \wv{cv_sift_vs_cosine_top_1_gain} [\wvci{cv_sift_vs_cosine_top_1_gain}] \\
    + DISK+LightGlue & \wv{cv_disk-lightglue_top_1} & \wv{cv_disk-lightglue_vs_cosine_top_1_gain} [\wvci{cv_disk-lightglue_vs_cosine_top_1_gain}] \\
    + LoFTR & \wv{cv_loftr_top_1} & \wv{cv_loftr_vs_cosine_top_1_gain} [\wvci{cv_loftr_vs_cosine_top_1_gain}] \\
    \bottomrule
  \end{tabular}
\end{table}

\paragraph{Learned verification helps; SIFT does not.}
Lookalike confusions suggest verifying the shortlist geometrically.
With blend weights chosen on held-out statues, both learned matchers improve unseen-photographer top-1; SIFT's gain is indistinguishable from zero (\cref{tab:verification}).
The step that matters is from hand-designed to learned matching; LoFTR's edge over DISK+LightGlue, \wv{cv_loftr_vs_disk-lightglue_top_1_gain} points (\wvci{cv_loftr_vs_disk-lightglue_top_1_gain}; McNemar $p = \wv{cv_loftr_vs_disk-lightglue_top_1_p_value}$), is marginal.
Unlike every other accuracy addition we measured, learned verification does not vanish on a stronger representation: on DINOv3's shortlists DISK+LightGlue still adds \wv{v3_cv_disk-lightglue_vs_cosine_top_1_gain} points (\wvci{v3_cv_disk-lightglue_vs_cosine_top_1_gain}), to \wv{v3_cv_disk-lightglue_top_1}\%, while SIFT again adds \wv{v3_cv_sift_vs_cosine_top_1_gain}.
Verification need not run on every query: verifying only the 30\% whose cosine top-two margin is smallest already reaches \wv{r2_cv_gate_disk-lightglue_30_top_1}\% with DISK+LightGlue and \wv{r2_cv_gate_loftr_30_top_1}\% with LoFTR.
The appendix gives pairwise comparisons, the same table for DINOv3, a top-50 shortlist, the weights each fold chose, the inlier-cap sensitivity, and example matches.

\paragraph{A margin lowers false acceptance.}
Retrieval always returns a nearest neighbour, so a photograph of a statue outside the reference set still gets a confident-looking answer~\cite{scheirer2013openset}.
Thresholding the top-1 cosine (\cref{sec:protocol}) separates known from unknown queries with an area under the ROC curve (AUROC) of \wv{dinov2_reject_cosine_auroc} for DINOv2 and \wv{dinov3_reject_cosine_auroc} for DINOv3 (FA@90 in \cref{tab:glance}).
A lookalike is one likely reason, and the margin between the best statue and the best \emph{other} statue measures that ambiguity: it raises AUROC to \wv{dinov2_reject_margin_auroc} and \wv{dinov3_reject_margin_auroc} and lowers FA@90 to \wv{dinov2_reject_margin_false_acceptance_90}\% and \wv{dinov3_reject_margin_false_acceptance_90}\%.

Known acceptance includes incorrect identifications: FA@90 alone does not establish a reliable operating point for deployment.

\paragraph{What does not help.}
Neither a linear probe nor class prototypes raise DINOv2's top-1, fusing DINOv2 with CLIP lowers its unseen-photographer top-1, query expansion lowers CLIP's unseen-photographer recall at ten, and registers do not raise DINOv2's.
A 3.5$\times$ larger DINOv2 is a partial exception: it raises unseen-photographer top-1 by \wv{null_dinov2-large_minus_dinov2_disjoint_r_at_1} points (\wvci{null_dinov2-large_minus_dinov2_disjoint_r_at_1}) but not recall at ten, where its interval rules out a gain above \wvhi{null_dinov2-large_minus_dinov2_disjoint_r_at_10} points, so it does not raise the re-ranking ceiling.
Ranking by local-feature geometry alone recovers only part of what cosine misses: of the 107 unseen-photographer queries whose statue is outside cosine's top ten, SIFT recovers 13 and DISK+LightGlue 33 to 36, at 1{,}690 matches per query.
Nor does a grid-pooled multi-vector representation, in the spirit of patch-wise retrieval~\cite{choi2026patchwise}, raise unseen-photographer top-1 for either backbone (appendix).
The appendix tabulates each change against its own baseline.

\section{Discussion}
\label{sec:discussion}

\paragraph{Where lookalikes matter, and where they do not.}
Lookalikes cause the most frequent confusions, and nearby statues are disproportionately confused, but most errors are diffuse, including statues lost in background and overlapping group labels.

\paragraph{The representation, and what is layered on it.}
Invariance-trained checkpoints lead on accuracy and rejection under our pooling rule (\cref{sec:retrieve}). Training data, architecture and preprocessing also differ, so this comparison does not isolate the pretraining objective.
Most additions help only a weaker representation: a linear probe adds 3.1 points to CLIP and nothing to DINOv2 (appendix).
Learned verification is the exception, adding \wv{v3_cv_disk-lightglue_vs_cosine_top_1_gain} points even to DINOv3 (\cref{sec:verify}).

\paragraph{What to expect in deployment.}
Leave-one-out accuracy is optimistic for a deployed guide, whose visitor is a new photographer.
In the Commons unseen-photographer arm DINOv2 scores 81.8\%, rising to 84.4\% with held-out learned verification, and DINOv3 scores 85.2\%, on statues documented by at least two people; the 125 others cannot be tested this way.
Cost is modest: exhaustive search takes 0.2\,ms a query on one CPU thread, and verifying a top-ten shortlist took 0.34\,s of wall clock with SIFT, 0.49\,s with DISK+LightGlue and 1.7\,s with LoFTR on one RTX 2080 Ti, which gating on the cosine margin cuts by over two-thirds.

\paragraph{Limitations.}
Every query here is a Commons upload; performance on casual street photographs remains unmeasured.
As queries, the 51 references taken on phones score 4.6 to 5.2 top-1 points below camera photographs (classified by EXIF camera model) for DINOv3, SigLIP~2 and DINOv2, and 15.6 below for CLIP, so the strongest backbones narrow the gap only slightly.
These estimates rest on 51 photographs without intervals; class composition and contributor behaviour may confound the camera comparison.
The benchmark was explored before the held-out re-ranking check, so that check guards only against weight selection, and other choices were made while looking at these queries.
DINOv3's margin accepts \wv{dinov3_reject_margin_false_acceptance_90}\% of unknown queries at 90\% known acceptance. This is calibrated on Commons uploads and does not measure correct-and-accepted identification on street photographs.

\section{Conclusion}
\label{sec:conclusion}

With no training, frozen DINOv2 features name the right statue first for \wv{dinov2_top_1}\% of leave-one-out queries, and the best of thirteen backbones from seven pretraining paradigms reaches \wv{fam_radio_top_1}\%.
Accuracy falls by about 0.8 points per doubling of the candidate pool.
Location narrowing helps, but less than an equal random pool does, and the penalty depends on the real layout; used as a weight instead of a filter, location lifts DINOv2 to \wv{dinov2_soft_prior_gps25_top_1}\% in the 25\,m simulation.
With the query's photographer withheld, DINOv2 reaches \wv{cv_cosine_top_1}\%, and a learned matcher with held-out weights raises it to as much as \wv{cv_loftr_top_1}\%, where SIFT adds nothing measurable.
At 90\% known acceptance, comparing the two best candidates cuts DINOv2's false acceptance from \wv{dinov2_reject_cosine_false_acceptance_90}\% to \wv{dinov2_reject_margin_false_acceptance_90}\% (DINOv3: \wv{dinov3_reject_cosine_false_acceptance_90}\% to \wv{dinov3_reject_margin_false_acceptance_90}\%); the gap to casual street photographs remains open.
Doppeldwarf-306 and its embeddings are published on \href{https://huggingface.co/datasets/turhancan97/doppeldwarf-306}{Hugging Face} and \href{https://www.kaggle.com/datasets/turhancankargin/wroclaw-dwarves}{Kaggle}; the code is on \href{https://github.com/turhancan97/doppeldwarf-306}{GitHub}.

\clearpage
\onecolumn
\appendix
\setcounter{figure}{0}
\setcounter{table}{0}
\renewcommand{\thefigure}{S\arabic{figure}}
\renewcommand{\thetable}{S\arabic{table}}
\renewcommand{\theHfigure}{S\arabic{figure}}
\renewcommand{\theHtable}{S\arabic{table}}
\renewcommand{\topfraction}{0.9}
\renewcommand{\bottomfraction}{0.7}
\renewcommand{\textfraction}{0.08}
\renewcommand{\floatpagefraction}{0.75}
\setcounter{topnumber}{4}
\setcounter{bottomnumber}{3}
\setcounter{totalnumber}{6}
\captionsetup[table]{position=top,skip=4pt}

\begin{center}
  {\Large\bfseries Appendix\par}
\end{center}

\section{Backbones and preprocessing}

\begin{table}[!htbp]
  \centering
  \scriptsize
  \setlength{\tabcolsep}{3pt}
  \caption{The thirteen frozen backbones, grouped by pretraining paradigm. Every checkpoint is pinned to an immutable revision that serves \texttt{safetensors}, and each uses the image processor released with it: DINOv2 resizes the short side to 256 and crops 224, C-RADIOv3 resizes the short side to 432 with no crop, as its model card does, and the rest produce 224-pixel inputs. Pooling is one rule fixed before any result: the model's released embedding where it has one, otherwise the mean of its patch tokens; MAE has a class token, but it is not a trained summary. C-RADIOv3's summary is three teacher-matched summaries concatenated. Width is the embedding dimension. Licences are those the Hub cards declare; CLIP's card states none, and the openai/CLIP repository is MIT.}
  \label{tab:roster}
  \begin{tabular}{@{}llllrrl@{}}
\toprule
Model & Checkpoint & Revision & Pooling & Input & Width & Licence \\
\midrule
\multicolumn{7}{@{}l}{\textit{Joint-embedding}} \\
\quad DINO B/16 & \texttt{timm/vit\_base\_patch16\_224.dino} & \texttt{4735e39} & CLS & 224 & 768 & Apache-2.0 \\
\quad DINOv2 B/14 & \texttt{facebook/dinov2-base} & \texttt{f9e44c8} & CLS & 224 & 768 & Apache-2.0 \\
\quad DINOv3 B/16 & \texttt{facebook/dinov3-vitb16-pretrain-lvd1689m} & \texttt{5931719} & CLS & 224 & 768 & DINOv3 Licence \\
\multicolumn{7}{@{}l}{\textit{JEPA}} \\
\quad I-JEPA H/14 & \texttt{facebook/ijepa\_vith14\_1k} & \texttt{f157467} & patch mean & 224 & 1280 & CC-BY-NC-4.0 \\
\multicolumn{7}{@{}l}{\textit{Masked image modelling}} \\
\quad MAE B/16 & \texttt{timm/vit\_base\_patch16\_224.mae} & \texttt{a64dffd} & patch mean & 224 & 768 & CC-BY-NC-4.0 \\
\quad BEiT B/16 & \texttt{microsoft/beit-base-patch16-224-pt22k} & \texttt{49128b4} & pooler (mean) & 224 & 768 & Apache-2.0 \\
\quad EVA-02 B/14 & \texttt{timm/eva02\_base\_patch14\_224.mim\_in22k} & \texttt{0ecfe3f} & patch mean & 224 & 768 & MIT \\
\multicolumn{7}{@{}l}{\textit{Vision-language}} \\
\quad CLIP B/32 & \texttt{openai/clip-vit-base-patch32} & \texttt{c237dc4} & projection & 224 & 512 & MIT \\
\quad CLIP B/16 & \texttt{openai/clip-vit-base-patch16} & \texttt{d815394} & projection & 224 & 512 & MIT \\
\quad SigLIP 2 B/16 & \texttt{google/siglip2-base-patch16-224} & \texttt{75de2d5} & attention pool & 224 & 768 & Apache-2.0 \\
\multicolumn{7}{@{}l}{\textit{Agglomerative}} \\
\quad C-RADIOv3 B/16 & \texttt{nvidia/C-RADIOv3-B} & \texttt{44653a0} & summary & 432 & 2304 & NVIDIA Open Model \\
\multicolumn{7}{@{}l}{\textit{Autoregressive}} \\
\quad AIMv2 L/14 & \texttt{timm/aimv2\_large\_patch14\_224.apple\_pt} & \texttt{4ecf1a8} & patch mean & 224 & 1024 & Apple ASCL \\
\multicolumn{7}{@{}l}{\textit{Supervised}} \\
\quad ViT B/16 (IN-21k) & \texttt{timm/vit\_base\_patch16\_224.augreg\_in21k} & \texttt{bb70896} & CLS & 224 & 768 & Apache-2.0 \\
\bottomrule
\end{tabular}

\end{table}

\begin{table}[!htbp]
  \centering
  \footnotesize
  \setlength{\tabcolsep}{3pt}
  \caption{All thirteen backbones by the measures of the main text, point estimates (\cref{fig:family} carries their intervals). Each query has one correct statue, so average precision equals reciprocal rank and MRR is the mAP. \emph{Unseen}: unseen-photographer top-1 and recall at ten over the 1{,}157 answerable queries. \emph{Ph.}: top-1 points attributable to the photographer, beyond the size-matched control. /2$\times$: top-1 points lost per doubling of the pool. \emph{Loc.}: geographic minus random top-1 at a pool of ten. AUROC: rejection by the margin score.}
  \label{tab:s-family-numbers}
  \begin{tabular}{@{}lrrrrrrrrr@{}}
\toprule
Model & Top-1 & Top-5 & MRR & Unseen & Unseen R@10 & Ph. & /2$\times$ & Loc. & AUROC \\
\midrule
\multicolumn{10}{@{}l}{\textit{Joint-embedding}} \\
\quad DINO B/16 & 92.0 & 95.3 & 0.936 & 74.4 & 86.3 & 6.9 & $-$0.9 & $-$0.8 & 0.946 \\
\quad DINOv2 B/14 & 93.1 & 95.7 & 0.943 & 81.8 & 90.8 & 2.6 & $-$0.8 & $-$0.7 & 0.927 \\
\quad DINOv3 B/16 & 94.4 & 96.9 & 0.956 & 85.2 & 92.8 & 1.7 & $-$0.7 & $-$1.1 & 0.959 \\
\multicolumn{10}{@{}l}{\textit{JEPA}} \\
\quad I-JEPA H/14 & 55.6 & 76.8 & 0.652 & 34.2 & 56.9 & 10.1 & $-$5.6 & $-$4.5 & 0.677 \\
\multicolumn{10}{@{}l}{\textit{Masked image modelling}} \\
\quad MAE B/16 & 46.8 & 63.5 & 0.546 & 14.3 & 32.1 & 14.8 & $-$6.3 & $-$2.8 & 0.616 \\
\quad BEiT B/16 & 52.8 & 70.1 & 0.612 & 20.7 & 42.9 & 15.0 & $-$5.6 & $-$4.1 & 0.667 \\
\quad EVA-02 B/14 & 84.5 & 92.1 & 0.880 & 63.2 & 80.4 & 8.6 & $-$1.9 & $-$1.9 & 0.864 \\
\multicolumn{10}{@{}l}{\textit{Vision-language}} \\
\quad CLIP B/32 & 82.9 & 90.7 & 0.866 & 54.1 & 75.1 & 13.2 & $-$2.1 & $-$2.1 & 0.852 \\
\quad CLIP B/16 & 87.7 & 92.6 & 0.900 & 63.6 & 79.9 & 10.0 & $-$1.5 & $-$1.1 & 0.895 \\
\quad SigLIP 2 B/16 & 91.2 & 94.7 & 0.929 & 75.0 & 86.7 & 6.2 & $-$1.0 & $-$1.0 & 0.916 \\
\multicolumn{10}{@{}l}{\textit{Agglomerative}} \\
\quad C-RADIOv3 B/16 & 94.7 & 96.5 & 0.956 & 84.3 & 91.8 & 3.3 & $-$0.6 & $-$0.4 & 0.947 \\
\multicolumn{10}{@{}l}{\textit{Autoregressive}} \\
\quad AIMv2 L/14 & 90.7 & 94.4 & 0.923 & 74.0 & 85.1 & 5.5 & $-$1.1 & $-$1.0 & 0.922 \\
\multicolumn{10}{@{}l}{\textit{Supervised}} \\
\quad ViT B/16 (IN-21k) & 89.4 & 93.7 & 0.914 & 62.5 & 78.6 & 11.7 & $-$1.3 & $-$1.7 & 0.917 \\
\bottomrule
\end{tabular}

\end{table}

DINOv2 and DINOv3 embeddings are the final-layer CLS token; for DINOv3 we verified that it equals the model's own pooled output, with the four register tokens following it. CLIP's is the projected image embedding. timm checkpoints are loaded strictly from the pinned weights file, so a missing or unexpected weight fails rather than being initialised. All embeddings are L2-normalised.

\begin{figure}[!htbp]
  \centering
  \includegraphics[width=17cm]{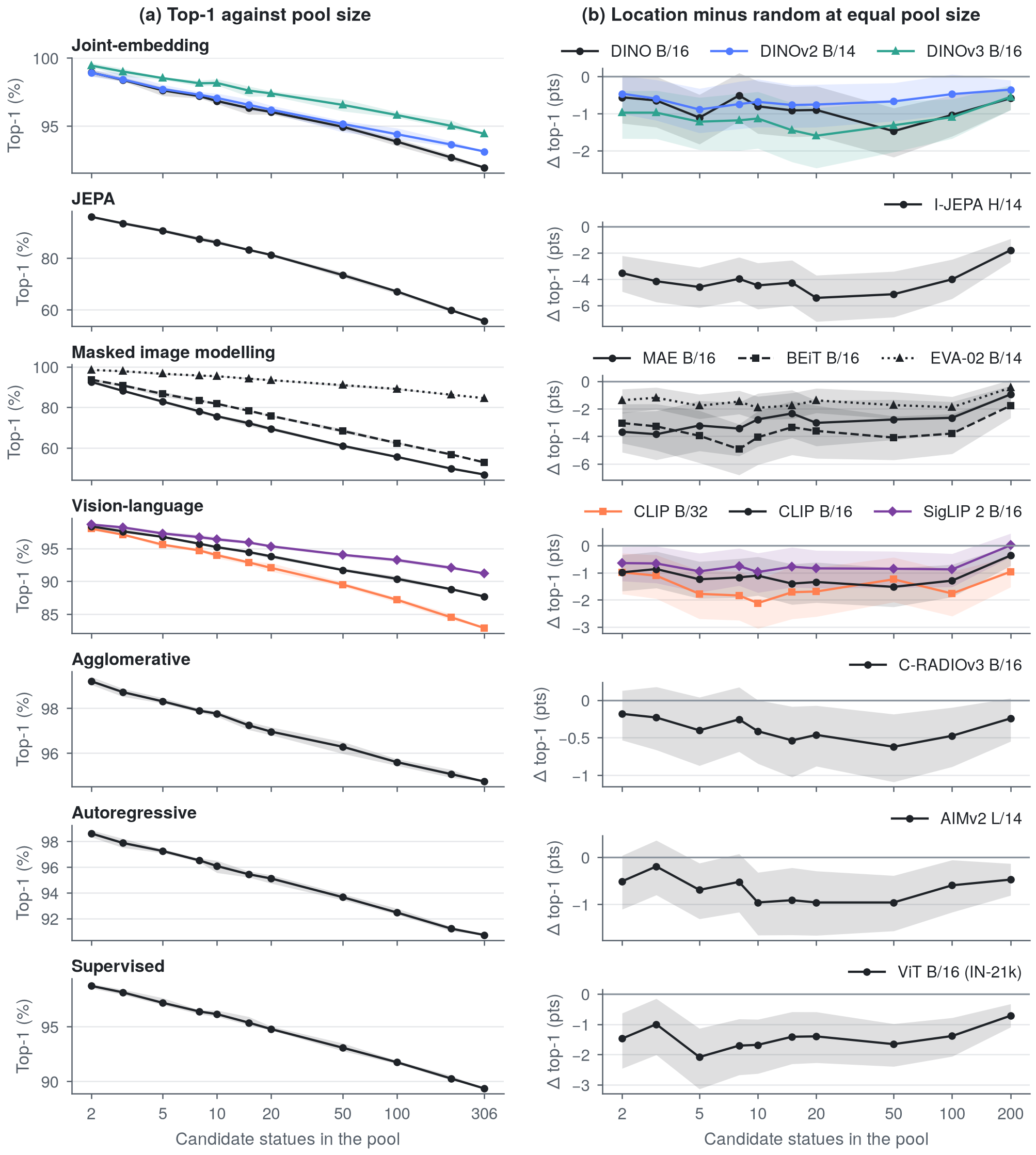}
  \caption{\textbf{All thirteen backbones, one row per pretraining paradigm.}
  (a)~Top-1 against the number of candidate statues, each query against its own statue plus $N-1$ random others; the band is the spread across five seeds.
  (b)~Top-1 of a geographic pool of the query statue's nearest neighbours minus an equal-sized random pool, over the 294 placed statues; the band is the 95\% interval from resampling whole statues.
  Each row keeps its own vertical scale, so the reconstruction-trained models' fall does not flatten the differences between the others.
  Each row names its paradigm on the left and its models on the right. The four backbones used throughout the paper keep their colours; the others are told apart by line style and marker.}
  \label{fig:s-family-curves}
\end{figure}

\section{Retrieval examples, top five}

\begin{figure}[p]
  \centering
  \includegraphics[width=13.6cm]{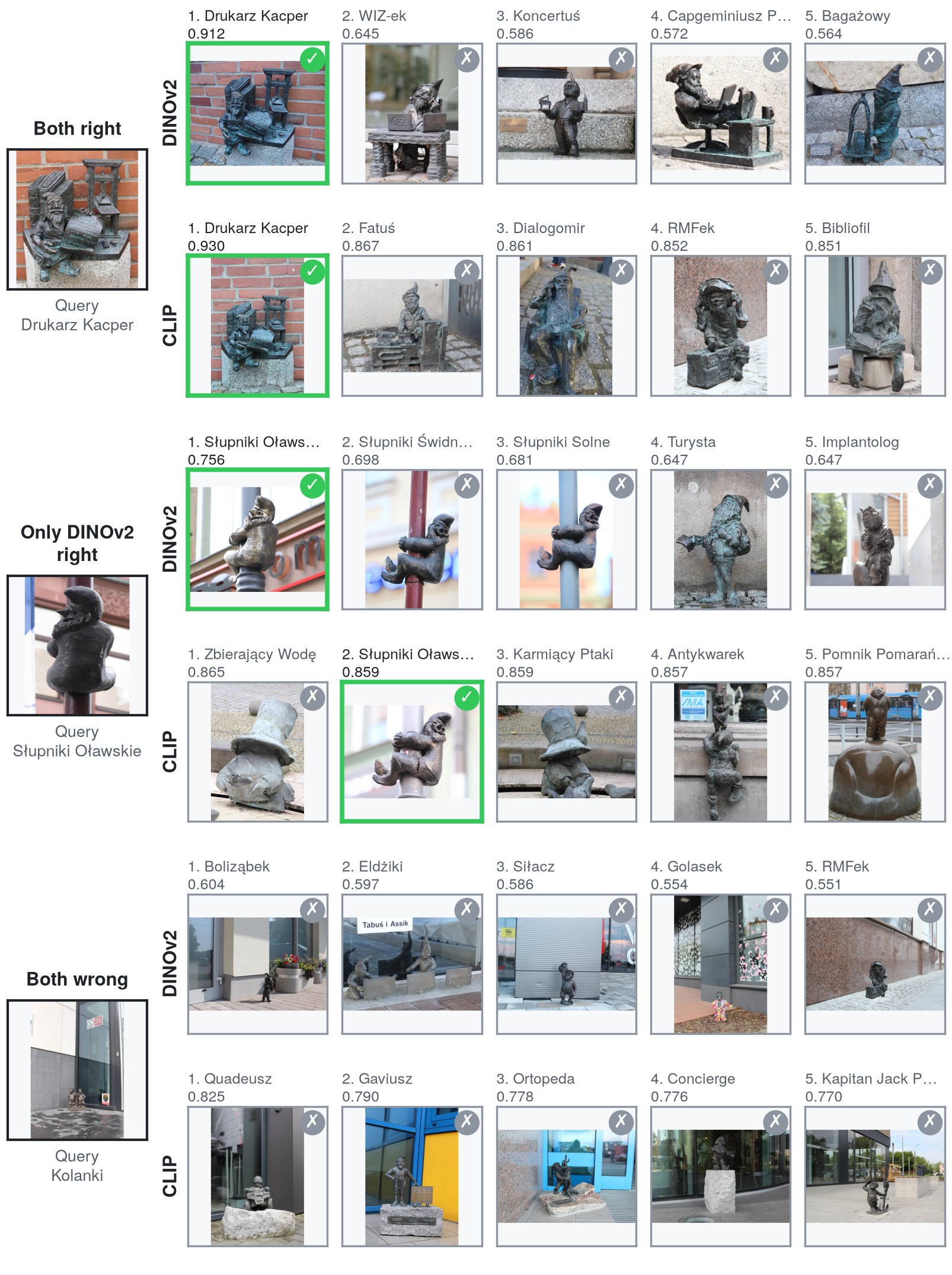}
  \caption{The main text's examples with the five statues each backbone ranks highest, with cosine similarity; a green frame and check mark is correct, a grey frame and cross wrong.}
  \label{fig:s-examples}
\end{figure}

\Cref{fig:s-examples} extends the main text's retrieval examples to the five statues each backbone ranks highest.

\section{Corpus and evaluation protocol}

\subsection{Corpus construction}

All 482 candidate Commons category-to-statue mappings carry human review decisions: 367 new, regular singular titles were approved in bulk by a naming rule, while plural and irregular titles were inspected individually; 469 were approved and 13 rejected, three of them approvals reversed after acquisition showed that their files duplicated another class. Images were admitted under public domain, CC0, CC~BY and CC~BY-SA only, with source, author, licence and licence URL required, a short side of at least 400 and a long side of at most 2{,}000 pixels. 36 files byte-identical across different labels (87 copies under 24 labels) were dropped from every label. A floor of three photographs per statue leaves 306 statues and 1{,}691 photographs by 122 photographers.

Two group classes coexist with their own members: \emph{Grajek i Meloman} (11 images) with \emph{Grajek} (6) and \emph{Meloman} (9), and \emph{Ogrodnik i Kierownik} (7) with \emph{Ogrodnik} (3), 36 of the 1{,}691 images in all. Each is its own class in every split; the main text reports what merging each group with its members changes.

Wikidata places 23 statues; 271 more are placed at the median of the camera positions recorded in their own photographs. On the 21 statues with both, the two disagree by a median of 9\,m and at most 147\,m. Twelve statues remain unplaced and are excluded from the geographic comparison only.

\subsection{The unseen-photographer control}

For each query, \emph{kept} is the set of references not taken by the query's photographer and \emph{own} the kept references of the correct statue; a query with no own reference is unanswerable and dropped, leaving 1{,}157 of 1{,}691. The control draws, per query and seed, as many correct-statue references as \emph{own} holds from all of that statue's references, and as many wrong-statue references as the rest of \emph{kept} holds from all wrong references. It therefore matches the total and the correct-statue reference counts while ignoring who took each photograph, but not the number of candidate statues: the withheld arm keeps a median of 252 per query against the control's 305, a smaller and so easier pool, which if anything understates the photographer's share.

Photographers are identified by Commons' author field compared exactly, so one person credited in two ways counts as two. To check for leakage through that gap, we took each answerable query's most similar kept reference of its own statue. Its DINOv2 cosine exceeds 0.95 for only 6 of the 1{,}157 queries, and all six are three pairs of files whose credits are spellings of one name or a crop of another contributor's photograph. Treating those credits as one person leaves unseen-photographer top-1 unchanged, at 81.85\% for DINOv2 and 85.22\% for DINOv3. Byte-identical files are already excluded across statue labels, and perceptual hashing finds no remaining near-duplicate between different statues.

\Cref{tab:s-strata} splits the unseen-photographer arm by how many references of the correct statue remain once the query's photographer is withheld. Accuracy rises steadily with that number for both backbones, so a large part of the drop from leave-one-out is scarcity of references rather than a change of photographer.

\begin{table}[!htbp]
  \centering
  \footnotesize
  \setlength{\tabcolsep}{3pt}
  \caption{Unseen-photographer top-1 (\%) by the number of references of the correct statue left after the query's photographer is withheld, with 95\% statue intervals. The four strata together are the 1{,}157 answerable queries.}
  \label{tab:s-strata}
  \begin{tabular}{@{}lrrcc@{}}
\toprule
References left & Queries & Statues & DINOv2 & DINOv3 \\
\midrule
1 & 207 & 66 & 60.4 [51.1, 70.1] & 69.6 [60.6, 78.9] \\
2 & 207 & 84 & 78.3 [71.1, 85.4] & 81.2 [74.8, 87.4] \\
3--5 & 347 & 99 & 85.0 [80.4, 89.6] & 88.2 [83.2, 92.4] \\
6+ & 396 & 67 & 92.2 [89.3, 94.9] & 92.9 [89.7, 95.7] \\
\bottomrule
\end{tabular}

\end{table}

\subsection{Uncertainty}

Main-text intervals resample whole statues 10{,}000 times with matching draws across compared arms, and report 2.5th and 97.5th percentiles; estimates stay per photograph. For comparison, the per-photograph Wilson intervals on full-pool top-1 are 91.8--94.3 (DINOv2), 81.0--84.6 (CLIP) and 93.2--95.4 (DINOv3), against statue intervals of \wvci{dinov2_top_1}, \wvci{clip_top_1} and \wvci{dinov3_top_1}. Seed spreads in the pool-size figure describe which random pools were drawn; statue intervals describe which statues were asked about. They answer different questions and are not combined.

\section{Re-ranking}
\label{sec:s-verify}

The blend weight is selected from the fixed grid $w \in \{0, .01, .02, .05, .1, .2, .3, .5, .75, 1, 1.5, 2\}$, with ties resolved in favour of the smaller weight.

\begin{table}[!htbp]
  \centering
  \footnotesize
  \setlength{\tabcolsep}{3pt}
  \caption{\textbf{Re-ranking with held-out weight selection}, unseen-photographer queries (1{,}157 queries, 181 statues).
  Each query's blend weight is chosen on the other four statue folds.
  W/L: queries won and lost against the row's baseline. The weight each fold chose is in \cref{tab:s-folds}.
  The top-50 row re-ranks fifty candidates instead of ten. The DINOv3 block repeats the protocol on DINOv3's top ten, over the same queries, and its gains are against DINOv3's cosine.}
  \label{tab:rerank}
  \begin{tabular}{@{}lccc@{}}
    \toprule
    Method & Top-1 & Gain [95\% statue CI] & W / L \\
    \midrule
    Cosine (DINOv2) & \wv{cv_cosine_top_1} & --- & --- \\
    + SIFT & \wv{cv_sift_top_1} & \wv{cv_sift_vs_cosine_top_1_gain} [\wvci{cv_sift_vs_cosine_top_1_gain}] & \wv{cv_sift_vs_cosine_top_1_wins} / \wv{cv_sift_vs_cosine_top_1_losses} \\
    + DISK+LightGlue & \wv{cv_disk-lightglue_top_1} & \wv{cv_disk-lightglue_vs_cosine_top_1_gain} [\wvci{cv_disk-lightglue_vs_cosine_top_1_gain}] & \wv{cv_disk-lightglue_vs_cosine_top_1_wins} / \wv{cv_disk-lightglue_vs_cosine_top_1_losses} \\
    + LoFTR & \wv{cv_loftr_top_1} & \wv{cv_loftr_vs_cosine_top_1_gain} [\wvci{cv_loftr_vs_cosine_top_1_gain}] & \wv{cv_loftr_vs_cosine_top_1_wins} / \wv{cv_loftr_vs_cosine_top_1_losses} \\
    + DISK+LG, top 50 & \wv{cv_k50_disk-lightglue_top_1} & \wv{cv_k50_disk-lightglue_vs_cosine_top_1_gain} [\wvci{cv_k50_disk-lightglue_vs_cosine_top_1_gain}] & \wv{cv_k50_disk-lightglue_vs_cosine_top_1_wins} / \wv{cv_k50_disk-lightglue_vs_cosine_top_1_losses} \\
    \midrule
    Cosine (DINOv3) & \wv{v3_cv_cosine_top_1} & --- & --- \\
    + SIFT & \wv{v3_cv_sift_top_1} & \wv{v3_cv_sift_vs_cosine_top_1_gain} [\wvci{v3_cv_sift_vs_cosine_top_1_gain}] & \wv{v3_cv_sift_vs_cosine_top_1_wins} / \wv{v3_cv_sift_vs_cosine_top_1_losses} \\
    + DISK+LightGlue & \wv{v3_cv_disk-lightglue_top_1} & \wv{v3_cv_disk-lightglue_vs_cosine_top_1_gain} [\wvci{v3_cv_disk-lightglue_vs_cosine_top_1_gain}] & \wv{v3_cv_disk-lightglue_vs_cosine_top_1_wins} / \wv{v3_cv_disk-lightglue_vs_cosine_top_1_losses} \\
    + LoFTR & \wv{v3_cv_loftr_top_1} & \wv{v3_cv_loftr_vs_cosine_top_1_gain} [\wvci{v3_cv_loftr_vs_cosine_top_1_gain}] & \wv{v3_cv_loftr_vs_cosine_top_1_wins} / \wv{v3_cv_loftr_vs_cosine_top_1_losses} \\
    \midrule
    DISK+LG vs SIFT & & \wv{cv_disk-lightglue_vs_sift_top_1_gain} [\wvci{cv_disk-lightglue_vs_sift_top_1_gain}] & \wv{cv_disk-lightglue_vs_sift_top_1_wins} / \wv{cv_disk-lightglue_vs_sift_top_1_losses} \\
    LoFTR vs DISK+LG & & \wv{cv_loftr_vs_disk-lightglue_top_1_gain} [\wvci{cv_loftr_vs_disk-lightglue_top_1_gain}] & \wv{cv_loftr_vs_disk-lightglue_top_1_wins} / \wv{cv_loftr_vs_disk-lightglue_top_1_losses} \\
    \bottomrule
  \end{tabular}
\end{table}

\begin{figure}[!htbp]
  \centering
  \begin{minipage}[t]{0.48\linewidth}
    \vspace{0pt}\centering
    \includegraphics[width=8.1cm]{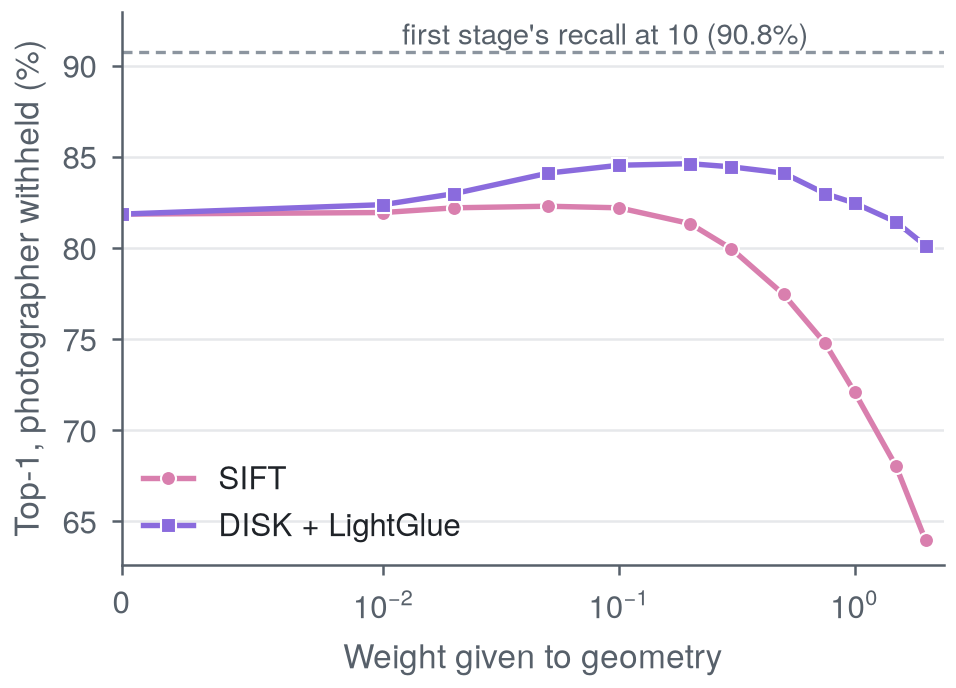}
    \captionof{figure}{The exploratory weight sweep, unseen-photographer queries, scored on the same queries that would choose a weight. Its peaks (SIFT 82.28\% at 0.05, DISK+LightGlue 84.62\% at 0.2) are in-sample; the main text reports held-out selection.}
    \label{fig:s-sweep}
  \end{minipage}\hfill
  \begin{minipage}[t]{0.48\linewidth}
    \vspace{0pt}\centering\footnotesize\setlength{\tabcolsep}{1.5pt}
    \captionof{table}{The blend weight each matcher selected on four statue folds, applied to the fifth. Selection top-1 is measured on the four.}
    \label{tab:s-folds}
    \begin{tabular}{@{}lccccc@{}}
\toprule
Matcher & Fold & Weight & Selection top-1 & Held-out queries & statues \\
\midrule
SIFT & 0 & 0.02 & 82.1 & 219 & 37 \\
 & 1 & 0.05 & 82.5 & 271 & 36 \\
 & 2 & 0.02 & 83.6 & 220 & 36 \\
 & 3 & 0.05 & 81.7 & 190 & 36 \\
 & 4 & 0.05 & 81.6 & 257 & 36 \\
\midrule
DISK+LightGlue & 0 & 0.1 & 84.6 & 219 & 37 \\
 & 1 & 0.2 & 85.3 & 271 & 36 \\
 & 2 & 0.2 & 85.8 & 220 & 36 \\
 & 3 & 0.2 & 83.9 & 190 & 36 \\
 & 4 & 0.2 & 83.7 & 257 & 36 \\
\midrule
LoFTR & 0 & 0.2 & 85.7 & 219 & 37 \\
 & 1 & 0.2 & 86.0 & 271 & 36 \\
 & 2 & 0.2 & 86.4 & 220 & 36 \\
 & 3 & 0.2 & 84.6 & 190 & 36 \\
 & 4 & 0.2 & 84.7 & 257 & 36 \\
\bottomrule
\end{tabular}

  \end{minipage}
\end{figure}

\begin{figure}[!htbp]
  \centering
  \includegraphics[width=17cm]{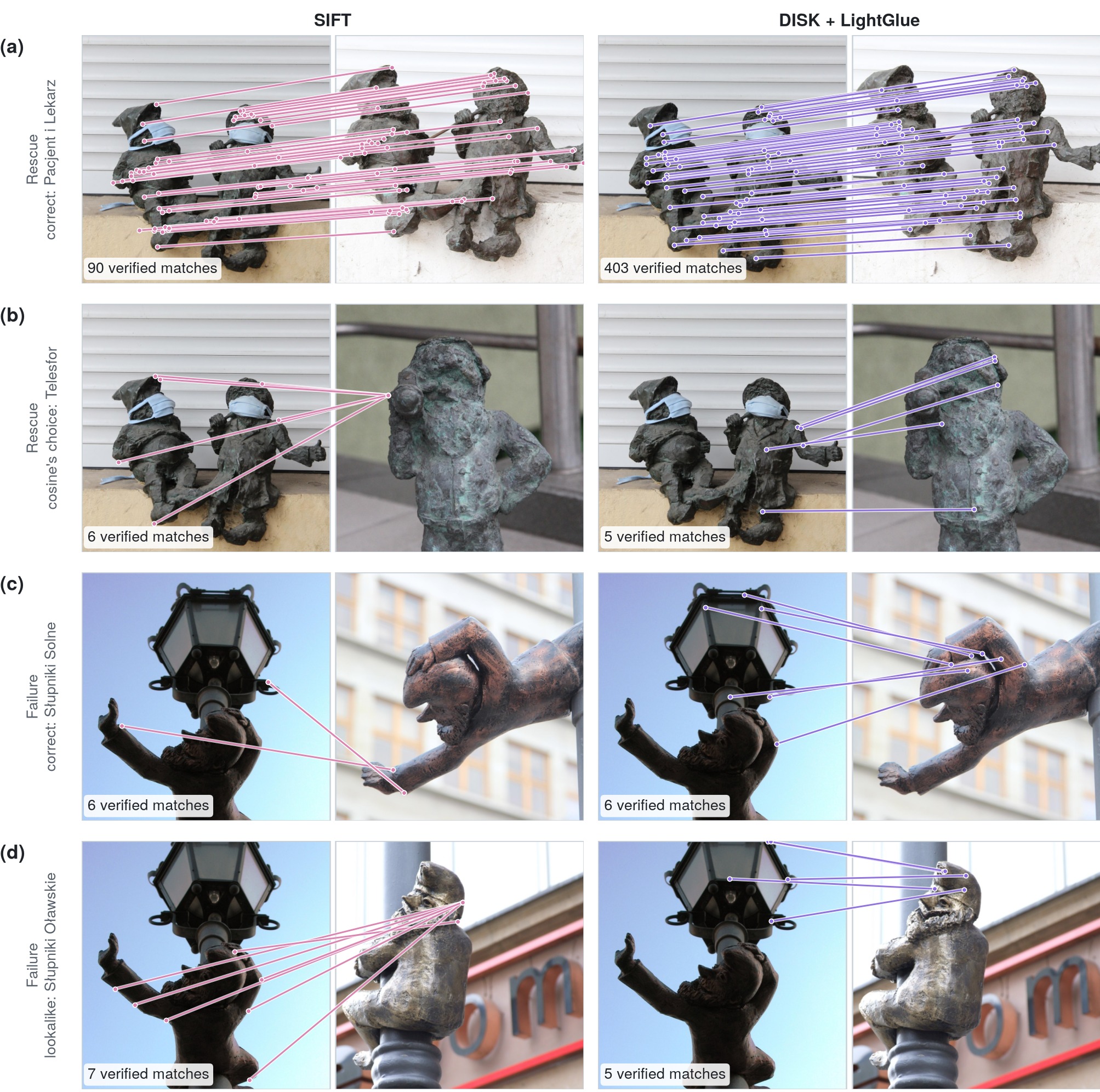}
  \caption{What the matchers verify, on two pairs of examples chosen by rule. Columns are matchers and rows are candidates; each panel shows the query (left) against one candidate statue (right), cropped square around the verified matches, and draws up to 60 of the matches that survive the homography check, while the count is all of them. \emph{Rescue}, rows (a) and (b): of the unseen-photographer queries whose correct statue cosine ranks below first, the one where DISK+LightGlue's journalled matches favour the correct statue over every other shortlisted statue by the widest margin, ties to the lower image ID. Each candidate is shown by the photograph re-ranking actually compared, its best-cosine reference by another photographer. Cosine prefers Telesfor. Both matchers verify the correct pair, SIFT with 90 matches and DISK+LightGlue with 403, although the statues wear face masks in the query and not in the reference, and both find only five or six on Telesfor. \emph{Failure}, rows (c) and (d): the teaser's query (\cref{fig:teaser}) against the best photograph of its own group of pillar dwarves and of the closest lookalike group. The query and its own group's photograph show different figures of that group, so they share little surface to match: neither matcher separates the two groups, and most matches land on the street lamp rather than a figure. Drawing re-runs each matcher, so a learned matcher's drawn count can differ from the journal's by one or two.}
  \label{fig:s-matches}
\end{figure}

Confusions between lookalikes suggest the classical remedy of verifying the shortlist geometrically (main text, protocol).
With the blend weight chosen on held-out statues, DISK+LightGlue raises unseen-photographer top-1 from 81.8\% to 84.4\%, a gain of 2.5 points whose statue interval excludes zero (\cref{tab:rerank}).
SIFT's gain is 0.3 points and indistinguishable from zero.
The detector-free LoFTR does best, at \wv{cv_loftr_top_1}\% (\wvci{cv_loftr_vs_cosine_top_1_gain} points over cosine).
Its edge over DISK+LightGlue is small, \wv{cv_loftr_vs_disk-lightglue_top_1_gain} points (\wvci{cv_loftr_vs_disk-lightglue_top_1_gain}; McNemar $p = \wv{cv_loftr_vs_disk-lightglue_top_1_p_value}$): the large step is from hand-designed to learned matching, not between learned matchers.
Verifying fifty candidates instead of ten lifts DISK+LightGlue from \wv{cv_disk-lightglue_top_1}\% to \wv{cv_k50_disk-lightglue_top_1}\%, so the shortlist, not only the matcher, limits verification.
The weights each fold selects are stable, which suggests the selection is not fitting noise.
Choosing the weight on the scored queries themselves would instead give 84.6\% and 82.3\% (\cref{fig:s-sweep}); the held-out figures are lower for both matchers. \Cref{fig:s-matches} shows, for one rescue and one failure chosen by rule, which matches each matcher verifies.

Over all 1{,}691 queries, the in-sample best SIFT weight lifts DINOv2 from 93.14\% to 94.03\% and CLIP from 82.91\% to 86.34\%. These full-pool sweeps have no journalled per-query evidence, so they could not be re-checked with held-out weights and are exploratory only.

Geometric evidence is weak in this regime.
Correct candidates have a median of 6 inliers against 4 for wrong ones once the photographer is withheld, and 58 against 4 when the photographer's own frames are available.
A blend can use a weak signal because it only reorders near-ties in cosine, and SIFT on specular, low-texture bronze evidently does not provide one.
\Cref{tab:s-cap} varies the inlier cap, re-choosing the weight on held-out statues each time. DISK+LightGlue's gain is \wv{sens_cap_disk_lightglue_min_gain} to \wv{sens_cap_disk_lightglue_max_gain} points across caps from 10 to 240 and LoFTR's \wv{sens_cap_loftr_min_gain} to \wv{sens_cap_loftr_max_gain}; only a cap of 10, which saturates before correct and wrong candidates separate, weakens them. SIFT's gain is not established at any cap.

\begin{table}[!htbp]
  \centering
  \footnotesize
  \setlength{\tabcolsep}{3pt}
  \caption{The inlier cap, DINOv2's top ten, unseen-photographer queries: held-out top-1 gain over cosine (points) for each cap on $n$ in $s + w\min(n, \mathrm{cap})/\mathrm{cap}$. The main text uses 30.}
  \label{tab:s-cap}
  \begin{tabular}{@{}lccc@{}}
\toprule
Inlier cap & SIFT & DISK+LightGlue & LoFTR \\
\midrule
10 & $-$0.2 [$-$0.9, 0.5] & 1.4 [$-$0.1, 2.9] & 0.2 [$-$0.3, 0.6] \\
20 & $-$0.1 [$-$0.9, 0.8] & 2.5 [1.1, 4.0] & 1.8 [0.6, 3.1] \\
30 & 0.3 [$-$0.4, 0.9] & 2.5 [1.3, 3.9] & 3.6 [2.2, 5.2] \\
60 & 0.6 [$-$0.1, 1.5] & 2.5 [1.2, 4.0] & 4.1 [2.7, 5.6] \\
120 & 0.6 [$-$0.1, 1.5] & 2.5 [1.1, 4.0] & 3.8 [2.3, 5.4] \\
240 & 0.6 [$-$0.1, 1.5] & 2.2 [0.6, 3.7] & 3.7 [2.1, 5.4] \\
\bottomrule
\end{tabular}

\end{table}

Verification can be spent where cosine is unsure. \Cref{tab:s-gated} verifies only the queries whose top-two cosine margin is smallest, using each fold's held-out weight; the order is fixed by cosine alone, so no matcher output selects the queries. Verifying 30\% of queries recovers all of DISK+LightGlue's gain and nearly all of LoFTR's.

\begin{table}[!htbp]
  \centering
  \footnotesize
  \setlength{\tabcolsep}{3pt}
  \caption{Margin-gated verification, DINOv2's top ten, unseen-photographer queries: top-1 (\%) when only the given fraction of queries, smallest cosine margin first, is re-ranked. 0\% is cosine and 100\% the held-out result of \cref{tab:rerank}. Each column draws its own bootstrap, so the identical 0\% rows differ slightly in their intervals.}
  \label{tab:s-gated}
  \begin{tabular}{@{}lccc@{}}
\toprule
Verified & SIFT & DISK+LightGlue & LoFTR \\
\midrule
0\% & 81.8 [78.2, 85.2] & 81.8 [78.3, 85.2] & 81.8 [78.2, 85.3] \\
5\% & 81.9 [78.4, 85.2] & 82.4 [78.8, 85.7] & 82.7 [79.1, 85.9] \\
10\% & 81.9 [78.4, 85.2] & 82.9 [79.3, 86.2] & 83.3 [79.9, 86.5] \\
20\% & 82.1 [78.5, 85.4] & 83.9 [80.4, 87.1] & 84.5 [81.2, 87.6] \\
30\% & 82.1 [78.6, 85.5] & 84.4 [81.1, 87.7] & 85.3 [82.1, 88.3] \\
50\% & 82.1 [78.6, 85.3] & 84.4 [80.7, 87.7] & 85.5 [82.2, 88.5] \\
75\% & 82.1 [78.6, 85.3] & 84.4 [80.9, 87.6] & 85.5 [82.2, 88.5] \\
100\% & 82.1 [78.5, 85.4] & 84.4 [80.8, 87.6] & 85.5 [82.2, 88.5] \\
\bottomrule
\end{tabular}

\end{table}

\Cref{tab:s-multiregion} tests a compact multi-vector alternative to verification. Each image keeps its CLS token and the mean of its final-layer patch tokens over each cell of a 2$\times$2 grid, or of a 2$\times$2 and a 3$\times$3 grid, every vector normalised; a query scores a reference by the mean over its vectors of the best-matching reference vector, and a statue by its best reference. It is training-free and in the spirit of patch-wise retrieval and region tokens, not a reproduction of either. On leave-one-out queries it lifts recall at ten slightly for DINOv2; once the photographer is withheld it helps neither backbone and lowers top-1, so whatever the regions add comes from the same photographer's frames rather than from the statue.

\begin{table}[!htbp]
  \centering
  \footnotesize
  \setlength{\tabcolsep}{3pt}
  \caption{The multi-region baseline: CLS cosine (\%, first row of each backbone) and each region set's paired difference from it (points), with 95\% statue intervals. \emph{Unseen}: the 1{,}157 answerable unseen-photographer queries.}
  \label{tab:s-multiregion}
  \begin{tabular}{@{}llcccc@{}}
\toprule
Backbone & Vectors & Top-1 & R@10 & Unseen top-1 & Unseen R@10 \\
\midrule
DINOv2 & CLS cosine & 93.1 [91.8, 94.4] & 96.2 [95.2, 97.1] & 81.8 [78.2, 85.2] & 90.8 [87.9, 93.3] \\
 & CLS + 2$\times$2 & 0.0 [$-$0.9, 0.9] & 0.7 [0.1, 1.3] & $-$2.0 [$-$4.4, 0.4] & 0.0 [$-$1.4, 1.4] \\
 & CLS + 2$\times$2 + 3$\times$3 & 0.7 [$-$0.3, 1.6] & 0.8 [0.1, 1.4] & $-$0.7 [$-$3.0, 1.6] & 0.5 [$-$1.0, 2.0] \\
DINOv3 & CLS cosine & 94.4 [93.1, 95.6] & 97.5 [96.7, 98.3] & 85.2 [81.9, 88.3] & 92.8 [90.4, 95.0] \\
 & CLS + 2$\times$2 & 0.1 [$-$0.7, 0.8] & 0.2 [$-$0.3, 0.7] & $-$1.9 [$-$4.1, 0.2] & $-$0.8 [$-$2.5, 0.8] \\
 & CLS + 2$\times$2 + 3$\times$3 & 0.0 [$-$0.8, 0.8] & $-$0.1 [$-$0.7, 0.5] & $-$2.6 [$-$5.2, $-$0.1] & $-$1.0 [$-$3.1, 0.9] \\
\bottomrule
\end{tabular}

\end{table}

Re-ranking has a ceiling: it reorders the first stage's top ten, and the correct statue is outside DINOv2's top ten for 107 of the 1{,}157 queries, capping any re-ranker at 90.75\%.

\section{Open-set rejection}
\label{sec:s-reject}

\begin{wrapfigure}{r}{8.4cm}
  \vspace{-1.2em}
  \centering
  \includegraphics[width=8.1cm]{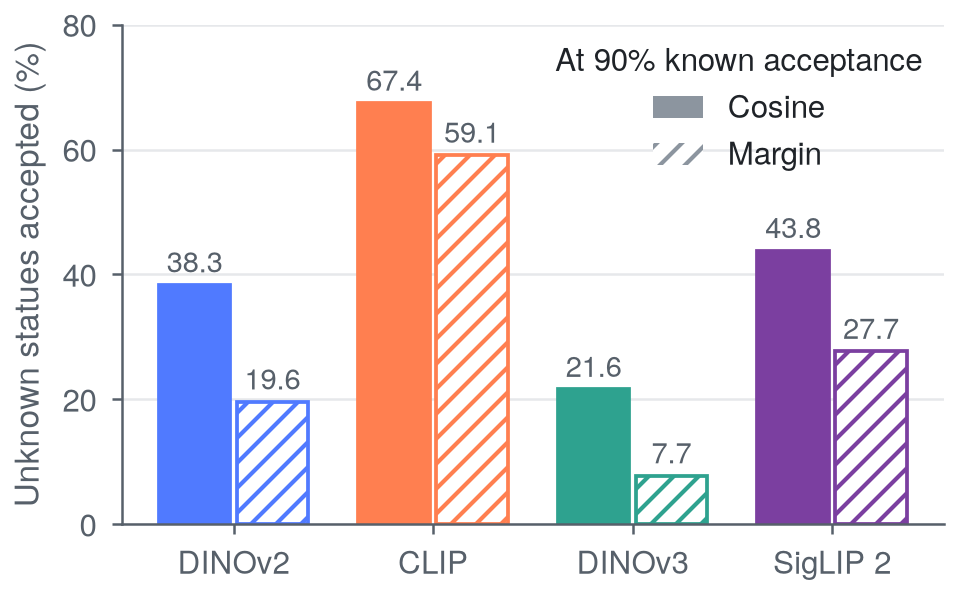}
  \caption{\textbf{A margin lowers false acceptance.}
  Unknown statues wrongly accepted when 90\% of known statues are accepted, with thresholds calibrated leave-one-class-out, for the top-1 cosine similarity (solid) and the margin between the best two statues (hatched).}
  \label{fig:reject}
  \vspace{-1em}
\end{wrapfigure}

Retrieval always returns a nearest neighbour, so a photograph of a statue outside the reference set still yields a confident-looking answer.
We threshold the top-1 cosine similarity, calibrated leave-one-class-out.
The test uses 1{,}691 known queries and 1{,}691 unknown ones, each made unknown by removing every photograph of its statue.
AUROC is 0.896 for DINOv2, 0.806 for CLIP and 0.926 for DINOv3.
Accepting 90\% of known statues admits 38.3\%, 67.4\% and 21.6\% of unknown ones respectively; even the best backbone would wrongly name roughly one unknown statue in five.
A lookalike is one likely reason, and the margin between the best statue and the best \emph{other} statue measures exactly that ambiguity.
Thresholding the margin instead (\cref{fig:reject}) raises AUROC for every backbone (DINOv2 \wv{dinov2_reject_margin_auroc}, a gain over cosine of \wv{dinov2_reject_margin_minus_cosine_auroc} with interval \wvci{dinov2_reject_margin_minus_cosine_auroc}; DINOv3 \wv{dinov3_reject_margin_auroc}) and roughly halves false acceptance at the same operating point for the self-supervised backbones: \wv{dinov2_reject_margin_false_acceptance_90}\% for DINOv2 and \wv{dinov3_reject_margin_false_acceptance_90}\% for DINOv3.
For DINOv3 that is about one unknown statue in thirteen. Known acceptance also includes incorrect identifications, so this operating point alone does not establish suitability for deployment.

The similarity to the fifth-nearest photograph is far worse than cosine (AUROC \wv{dinov2_reject_knn_auroc}), because with a median of four photographs per statue the fifth neighbour is usually another statue.
Geometry does not close the gap for DINOv2.
Blending inliers into the score raises AUROC by 0.011 with the photographer's frames available.
With the photographer withheld, thresholding the largest inlier count over the top ten reaches an AUROC of 0.707, below cosine's 0.798.

\begin{table}[!htbp]
  \centering
  \footnotesize
  \setlength{\tabcolsep}{3pt}
  \caption{Rejection signals. AUROC separating known from unknown statues, and false acceptance of unknown statues (\%) at 90\% known acceptance, calibrated leave-one-class-out. \emph{Margin}: top-1 similarity minus the best similarity of a different statue. $k$-NN: similarity to the fifth nearest photograph.}
  \label{tab:s-rejection}
  \begin{tabular}{@{}lcccccc@{}}
\toprule
& \multicolumn{2}{c}{Cosine} & \multicolumn{2}{c}{Margin} & \multicolumn{2}{c}{$k$-NN} \\
Backbone & AUROC & FA@90 & AUROC & FA@90 & AUROC & FA@90 \\
\midrule
DINOv2 & 0.896 [0.883, 0.909] & 38.3 & 0.927 [0.914, 0.940] & 19.6 & 0.637 [0.611, 0.665] & 82.2 \\
CLIP & 0.806 [0.788, 0.823] & 67.4 & 0.852 [0.834, 0.869] & 59.1 & 0.571 [0.559, 0.585] & 87.0 \\
DINOv3 & 0.926 [0.914, 0.937] & 21.6 & 0.959 [0.949, 0.967] & 7.7 & 0.660 [0.633, 0.689] & 84.0 \\
SigLIP~2 & 0.877 [0.861, 0.892] & 43.8 & 0.916 [0.902, 0.929] & 27.7 & 0.630 [0.607, 0.655] & 83.1 \\
\bottomrule
\end{tabular}

\end{table}

\section{Other changes to the first stage}

\begin{table}[!htbp]
  \centering
  \footnotesize
  \setlength{\tabcolsep}{2.5pt}
  \caption{\textbf{Changes to the first stage}, in accuracy points against the matching DINOv2 (or CLIP) baseline.
  R@10 is recall at ten, the re-ranking ceiling.
  ViT-L: paired, 62 won and 32 lost at top-1 ($p = 0.003$) but 22 and 20 at R@10 ($p = 0.88$); its R@10 statue interval, \wvci{null_dinov2-large_minus_dinov2_disjoint_r_at_10}, bounds any hidden gain.
  The rescue range counts tied inlier scores for and against the correct statue.}
  \label{tab:negative}
  \begin{tabular}{@{}llr@{}}
    \toprule
    Change (DINOv2 unless noted) & Measured as & Effect \\
    \midrule
    Linear probe (CLIP: $+3.1$) & top-1, all & $-0.1$ \\
    Class prototypes & top-1, all & $-8.0$ \\
    ViT-L/14, 3.5$\times$ parameters & top-1, unseen ph. & $+2.6$ \\
    & R@10, unseen ph. & $+0.2$ \\
    Registers & R@10, unseen ph. & $-1.3$ \\
    DINOv3 ViT-B/16 & R@10, unseen ph. & $+2.1$ \\
    Fusion with CLIP & top-1, unseen ph. & $-2.3$ \\
    Query expansion (CLIP) & R@10, unseen ph. & $-6.9$ \\
    SIFT as first stage & R@10, unseen ph. & $-47.3$ \\
    \multicolumn{3}{@{}l}{\emph{Of the 107 queries cosine misses at ten, recovered into the top ten by:}} \\
    \quad SIFT, geometry only & rescued & 13 \\
    \quad DISK+LightGlue, geometry only & rescued & 33--36 \\
    \bottomrule
  \end{tabular}
\end{table}

\Cref{tab:negative} collects the other changes we measured, scoped to the metric each was tested on.
A linear probe trained on the frozen embeddings changes DINOv2 by $-0.1$ points and CLIP by $+3.1$.
Scaling DINOv2 to ViT-L/14 improves unseen-photographer top-1 by 2.6 points ($p = 0.003$) but leaves recall at ten essentially unchanged ($+0.2$, $p = 0.88$), so it does not raise the re-ranking ceiling.
A null result is only as strong as the effect it could have seen: the statue interval rules out a gain in recall at ten larger than \wvhi{null_dinov2-large_minus_dinov2_disjoint_r_at_10} points.
Registers do not help either: \wv{null_dinov2-registers_minus_dinov2_disjoint_r_at_10} points (\wvci{null_dinov2-registers_minus_dinov2_disjoint_r_at_10}), an interval with no room for a gain.
DINOv3, a newer pretraining at the base size, raises that recall by 2.1 points to 92.8\%.
Fusing DINOv2 with CLIP lowers DINOv2's unseen-photographer top-1 by 2.3 points.
Query expansion assumes the top results are mostly right; it lowers CLIP's recall at ten by 6.9 points, consistent with the expanded query drifting toward lookalikes.
Ranking every reference by SIFT inlier count alone, with no appearance, finds the correct statue within ten for 43.5\% of unseen-photographer queries against 90.8\% for cosine, and rescues 13 of the 107 queries cosine misses.
The same ranking with DISK+LightGlue rescues 33 to 36 of them, depending on how tied inlier counts are broken.
That is a partial recovery, about a third, measured on the misses alone, at 1{,}690 matches per query, and leaves two-thirds of them lost.

The tables below give the measurements behind the classifier, capacity and geometry rows of \cref{tab:negative}, and the camera-origin bound the main text's limitations rest on. In them, \emph{photographer-disjoint} and \emph{photographer withheld} denote the main text's unseen-photographer arm.

\begin{table}[ht]
\caption{Trained classifiers against retrieval, on identical folds. A linear probe is fitted per fold on the same frozen embeddings; class prototypes average each statue's reference vectors into one centroid. Probes are regularised weakly ($C = 100$): the embeddings are L2-normalised, so a conventional $C = 1$ underfits badly.}\label{si:probe}
\begin{tabular}{@{}lcccccc@{}}
\toprule
Method & DINOv2 top-1 & top-5 & MRR & CLIP top-1 & top-5 & MRR \\
\midrule
Cosine retrieval & 93.14 & 95.68 & 0.943 & 82.91 & 90.72 & 0.866 \\
Linear probe & 93.02 & 95.56 & 0.942 & 86.04 & 93.14 & 0.891 \\
Class prototypes & 85.10 & 90.12 & 0.876 & 74.57 & 86.10 & 0.797 \\
\botrule
\end{tabular}
\end{table}

\begin{table}[ht]
\caption{Capacity does not lift the ceiling. Photographer-disjoint recall for DINOv2 ViT-B/14 against three larger or register-equipped checkpoints, with the change in parentheses. Paired against ViT-B/14, wins/losses and exact McNemar $p$: $k=1$: 62/32, $p = 0.003$; $k=5$: 28/21, $p = 0.392$; $k=10$: 22/20, $p = 0.878$; $k=20$: 19/15, $p = 0.608$; $k=50$: 14/20, $p = 0.392$; $k=100$: 10/21, $p = 0.071$. Six cutoffs are compared here, so a Bonferroni-corrected threshold is $0.05/6 = 0.008$; only $k = 1$ clears it, and the cutoff that matters for re-ranking, $k = 10$, is a 0.17-point move at $p = 0.878$.}\label{si:capacity}
\begin{tabular}{@{}lcccc@{}}
\toprule
$k$ & ViT-B/14 & DINOv2 ViT-L/14 & DINOv2 + registers & DINOv2 ViT-L/14 + registers \\
\midrule
1 & 81.85 & 84.44 (+2.59) & 80.55 (−1.30) & 80.81 (−1.04) \\
5 & 89.02 & 89.63 (+0.61) & 87.64 (−1.38) & 86.95 (−2.07) \\
10 & 90.75 & 90.92 (+0.17) & 89.46 (−1.30) & 88.94 (−1.82) \\
20 & 92.05 & 92.39 (+0.35) & 91.10 (−0.95) & 90.75 (−1.30) \\
50 & 94.21 & 93.69 (−0.52) & 94.04 (−0.17) & 94.38 (+0.17) \\
100 & 96.20 & 95.25 (−0.95) & 96.02 (−0.17) & 95.76 (−0.43) \\
\botrule
\end{tabular}
\end{table}

\begin{table}[ht]
\caption{Local matching as a first stage rather than a re-ranker: every one of the 306 candidates ranked by SIFT inlier count alone, no embedding involved. Inlier counts are integers and tie heavily, so the rank of the correct statue is reported as the best and worst positions consistent with its score rather than as whichever order the manifest happened to impose. The rescue rate is the share of the statues appearance misses that geometry finds. It retrieves far worse than appearance and rescues a minority of its failures.}\label{si:geometryfirst}
\begin{tabular}{@{}llccccc@{}}
\toprule
Arm & $k$ & Appearance & Geometry (best) & (worst) & Misses & Rescued \\
\midrule
All references & 1 & 94.21 & 70.70 & 69.75 & 67 & 32.8 \\
 & 5 & 96.97 & 76.49 & 75.02 & 35 & 31.4 \\
 & 10 & 97.49 & 79.60 & 76.66 & 29 & 37.9 \\
Photographer withheld & 1 & 81.85 & 28.35 & 26.97 & 210 & 8.1 \\
 & 5 & 89.02 & 38.20 & 34.75 & 127 & 9.4 \\
 & 10 & 90.75 & 43.47 & 39.07 & 107 & 12.1 \\
\botrule
\end{tabular}
\end{table}

\begin{table}[ht]
\caption{Queries by the capture device recorded in the photograph's metadata. The median reference count is given because it is the obvious confound and it does not differ meaningfully between conditions. Fifty-one phone queries make that row imprecise, and a phone upload to a photo commons is a deliberate photograph rather than a snapshot, so the gap is a lower bound on what a street query costs.}\label{si:camera}
\begin{tabular}{@{}lrrcccc@{}}
\toprule
Device & Queries & Refs & DINOv2 top-1 & top-5 & CLIP top-1 & top-5 \\
\midrule
Camera & 1565 & 6 & 93.48 & 95.97 & 84.22 & 91.82 \\
Phone & 51 & 7 & 88.24 & 90.20 & 68.63 & 76.47 \\
Unrecorded & 75 & 7 & 89.33 & 93.33 & 65.33 & 77.33 \\
\botrule
\end{tabular}
\end{table}

The same analysis on the two newer backbones puts phone queries \wv{cg3_top_1_gap} points below camera queries for DINOv3 and \wv{cgs_top_1_gap} for SigLIP~2, the other figures the main text's limitations quote.

\section{Tests of the explanations}

Every table below is generated from the validation artifact; each comparison carries a 95\% statue-bootstrap interval.

\begin{table}[!htbp]
  \centering
  \footnotesize
  \setlength{\tabcolsep}{4pt}
  \caption{Lookalike families and errors. Two statues are linked when each is among the other's three nearest by mean embedding under at least two backbones, and families are the connected groups (main text, protocol); each backbone is judged against families drawn by the \emph{other} three, never from its own embedding. \emph{In a family}: statues in such a family and their share of the 306. \emph{Within a family}: share of the backbone's errors that name another member of the query's own family. \emph{Error rate}: share of queries answered wrongly, for statues inside and outside a family. Brackets are 95\% statue intervals.}
  \label{tab:s-families}
  \begin{tabular}{@{}lccccc@{}}
    \toprule
    & & & Errors within & \multicolumn{2}{c}{Error rate (\%)} \\
    \cmidrule(l){5-6}
    Backbone & In a family & Errors & a family (\%) & inside & outside \\
    \midrule
    DINOv2 & \wv{dinov2_families_statues} (\wv{dinov2_families_statue_share}\%) & \wv{dinov2_families_errors} & \wv{dinov2_families_error_share} [\wvci{dinov2_families_error_share}] & \wv{dinov2_families_inside_error_rate} [\wvci{dinov2_families_inside_error_rate}] & \wv{dinov2_families_outside_error_rate} [\wvci{dinov2_families_outside_error_rate}] \\
    CLIP & \wv{clip_families_statues} (\wv{clip_families_statue_share}\%) & \wv{clip_families_errors} & \wv{clip_families_error_share} [\wvci{clip_families_error_share}] & \wv{clip_families_inside_error_rate} [\wvci{clip_families_inside_error_rate}] & \wv{clip_families_outside_error_rate} [\wvci{clip_families_outside_error_rate}] \\
    DINOv3 & \wv{dinov3_families_statues} (\wv{dinov3_families_statue_share}\%) & \wv{dinov3_families_errors} & \wv{dinov3_families_error_share} [\wvci{dinov3_families_error_share}] & \wv{dinov3_families_inside_error_rate} [\wvci{dinov3_families_inside_error_rate}] & \wv{dinov3_families_outside_error_rate} [\wvci{dinov3_families_outside_error_rate}] \\
    SigLIP~2 & \wv{siglip2_families_statues} (\wv{siglip2_families_statue_share}\%) & \wv{siglip2_families_errors} & \wv{siglip2_families_error_share} [\wvci{siglip2_families_error_share}] & \wv{siglip2_families_inside_error_rate} [\wvci{siglip2_families_inside_error_rate}] & \wv{siglip2_families_outside_error_rate} [\wvci{siglip2_families_outside_error_rate}] \\
    \bottomrule
  \end{tabular}
\end{table}

\begin{table}[!htbp]
  \centering
  \footnotesize
  \setlength{\tabcolsep}{3pt}
  \caption{Permutation test of co-location. \emph{Real}: geographic minus random top-1 (points) with statues where they stand. \emph{Shuffled}: the same with statue positions permuted among the 294 placed statues, mean over 200 permutations with the central 95\% of permutations. $p$: share of permutations at least as harmful as the real layout, computed on the geographic-minus-random difference averaged over the three pool sizes, counting the real layout.}
  \label{tab:s-permutation}
  \begin{tabular}{@{}lccccccc@{}}
\toprule
 & \multicolumn{2}{c}{Pool 5} & \multicolumn{2}{c}{Pool 10} & \multicolumn{2}{c}{Pool 20} &  \\
Backbone & Real & Shuffled & Real & Shuffled & Real & Shuffled & $p$ \\
\midrule
DINOv2 & $-$0.9 [$-$1.5, $-$0.3] & $-$0.1 [$-$0.5, 0.3] & $-$0.7 [$-$1.4, $-$0.1] & 0.2 [$-$0.3, 0.6] & $-$0.8 [$-$1.4, $-$0.2] & 0.0 [$-$0.4, 0.5] & 0.005 \\
CLIP & $-$1.8 [$-$2.7, $-$0.9] & 0.0 [$-$0.6, 0.6] & $-$2.1 [$-$3.1, $-$1.2] & $-$0.1 [$-$0.7, 0.6] & $-$1.7 [$-$2.6, $-$0.8] & 0.2 [$-$0.5, 0.9] & 0.005 \\
DINOv3 & $-$1.2 [$-$2.0, $-$0.6] & $-$0.1 [$-$0.4, 0.3] & $-$1.1 [$-$1.9, $-$0.4] & 0.1 [$-$0.2, 0.5] & $-$1.6 [$-$2.5, $-$0.8] & $-$0.1 [$-$0.6, 0.2] & 0.005 \\
SigLIP~2 & $-$1.0 [$-$1.7, $-$0.3] & 0.0 [$-$0.4, 0.5] & $-$1.0 [$-$1.7, $-$0.3] & $-$0.1 [$-$0.6, 0.3] & $-$0.8 [$-$1.6, $-$0.2] & 0.1 [$-$0.4, 0.5] & 0.005 \\
\bottomrule
\end{tabular}

\end{table}

\begin{table}[!htbp]
  \centering
  \footnotesize
  \setlength{\tabcolsep}{3pt}
  \caption{Sensitivity to coordinate error, DINOv2: geographic minus random top-1 (points) when every statue is moved the given distance in a random direction, averaged over five draws.}
  \label{tab:s-noise}
  \begin{tabular}{@{}lcccc@{}}
\toprule
Pool & As placed & 9\,m & 50\,m & 150\,m \\
\midrule
2 & $-$0.5 [$-$1.0, 0.0] & $-$0.5 [$-$0.9, 0.0] & $-$0.3 [$-$0.7, 0.0] & $-$0.2 [$-$0.6, 0.1] \\
3 & $-$0.6 [$-$1.2, $-$0.1] & $-$0.5 [$-$1.0, $-$0.1] & $-$0.3 [$-$0.7, 0.0] & $-$0.3 [$-$0.6, 0.0] \\
5 & $-$0.9 [$-$1.5, $-$0.3] & $-$0.9 [$-$1.5, $-$0.3] & $-$0.6 [$-$1.1, $-$0.2] & $-$0.6 [$-$1.1, $-$0.2] \\
8 & $-$0.7 [$-$1.4, $-$0.1] & $-$0.8 [$-$1.5, $-$0.2] & $-$0.8 [$-$1.3, $-$0.3] & $-$0.5 [$-$1.0, $-$0.1] \\
10 & $-$0.7 [$-$1.4, $-$0.1] & $-$0.7 [$-$1.4, $-$0.1] & $-$0.6 [$-$1.2, $-$0.1] & $-$0.4 [$-$0.9, 0.0] \\
15 & $-$0.8 [$-$1.4, $-$0.2] & $-$0.7 [$-$1.3, $-$0.1] & $-$0.7 [$-$1.2, $-$0.2] & $-$0.4 [$-$0.8, 0.0] \\
20 & $-$0.8 [$-$1.4, $-$0.2] & $-$0.8 [$-$1.4, $-$0.2] & $-$0.8 [$-$1.4, $-$0.3] & $-$0.6 [$-$1.1, $-$0.2] \\
50 & $-$0.7 [$-$1.2, $-$0.2] & $-$0.7 [$-$1.3, $-$0.2] & $-$0.6 [$-$1.2, $-$0.1] & $-$0.6 [$-$1.1, $-$0.1] \\
100 & $-$0.5 [$-$1.0, 0.1] & $-$0.5 [$-$1.0, 0.1] & $-$0.5 [$-$1.0, 0.0] & $-$0.5 [$-$1.1, $-$0.1] \\
200 & $-$0.4 [$-$0.6, $-$0.1] & $-$0.4 [$-$0.6, $-$0.1] & $-$0.4 [$-$0.6, $-$0.1] & $-$0.4 [$-$0.6, $-$0.1] \\
\bottomrule
\end{tabular}

\end{table}

\begin{table}[!htbp]
  \centering
  \footnotesize
  \setlength{\tabcolsep}{3pt}
  \caption{Location as a weight, DINOv2: top-1 (\%) over the 294 placed statues for each prior weight $\lambda$ and distance between user and statue. The rows above the rule are in-sample and descriptive; \emph{Held out} chooses $\lambda$ on four statue folds and scores the fifth, and \emph{Chosen} lists the weight each fold selected.}
  \label{tab:s-soft-prior}
  \begin{tabular}{@{}lccc@{}}
\toprule
$\lambda$ & 10\,m & 25\,m & 50\,m \\
\midrule
0 & 93.4 & 93.4 & 93.4 \\
0.005 & 93.6 & 93.5 & 93.5 \\
0.01 & 93.7 & 93.7 & 93.7 \\
0.02 & 94.0 & 94.0 & 93.9 \\
0.05 & 94.8 & 94.7 & 94.5 \\
0.1 & 95.7 & 95.6 & 95.4 \\
0.2 & 96.8 & 96.6 & 96.4 \\
0.5 & 98.4 & 98.1 & 97.5 \\
1 & 98.9 & 98.7 & 97.3 \\
2 & 99.1 & 98.3 & 92.1 \\
5 & 99.0 & 93.1 & 75.9 \\
10 & 98.0 & 85.1 & 68.1 \\
20 & 95.5 & 80.1 & 64.0 \\
\midrule
Held out & 99.1 [98.6, 99.6] & 98.7 [98.0, 99.3] & 97.5 [96.7, 98.2] \\
Chosen $\lambda$ & 2, 2, 2, 2, 2 & 1, 1, 1, 1, 1 & 0.5, 0.5, 0.5, 0.5, 0.5 \\
\bottomrule
\end{tabular}

\end{table}

\begin{table}[!htbp]
  \centering
  \footnotesize
  \setlength{\tabcolsep}{3pt}
  \caption{The prior's kernel, DINOv2: held-out top-1 (\%) over the 294 placed statues for each kernel shape and scale $s$, at each distance between user and statue. $\lambda$ is chosen on held-out statue folds in every row; the exponential at 200\,m is the main text's. The gain shrinks as the user moves farther from the statue, and a wider kernel then holds more of it.}
  \label{tab:s-kernel}
  \begin{tabular}{@{}llcccc@{}}
\toprule
Kernel & Scale $s$ & 10\,m & 25\,m & 50\,m & 100\,m \\
\midrule
No prior & --- & 93.4 [92.1, 94.6] & 93.4 [92.1, 94.6] & 93.4 [92.1, 94.6] & 93.4 [92.1, 94.6] \\
$e^{-d/s}$ & 50\,m & 99.1 [98.5, 99.6] & 97.7 [96.9, 98.4] & 95.3 [94.2, 96.4] & 93.7 [92.5, 94.9] \\
 & 100\,m & 99.1 [98.5, 99.6] & 98.3 [97.6, 99.0] & 96.8 [95.9, 97.6] & 94.9 [93.7, 96.0] \\
 & 200\,m & 99.1 [98.6, 99.6] & 98.7 [98.0, 99.3] & 97.5 [96.7, 98.2] & 95.4 [94.2, 96.4] \\
 & 400\,m & 99.1 [98.6, 99.6] & 98.7 [98.0, 99.3] & 98.0 [97.3, 98.7] & 96.6 [95.6, 97.5] \\
 & 800\,m & 99.1 [98.6, 99.6] & 98.6 [97.9, 99.2] & 98.1 [97.3, 98.8] & 97.0 [96.1, 97.8] \\
$e^{-d^2/2s^2}$ & 50\,m & 99.2 [98.6, 99.6] & 98.7 [98.1, 99.3] & 96.2 [95.2, 97.2] & 93.6 [92.3, 94.8] \\
 & 100\,m & 99.1 [98.6, 99.6] & 98.8 [98.1, 99.4] & 97.9 [97.1, 98.6] & 95.5 [94.4, 96.6] \\
 & 200\,m & 99.2 [98.6, 99.7] & 98.7 [98.1, 99.3] & 98.3 [97.6, 99.0] & 97.1 [96.1, 97.9] \\
 & 400\,m & 98.9 [98.2, 99.4] & 98.8 [98.1, 99.3] & 98.3 [97.6, 99.0] & 97.5 [96.6, 98.2] \\
 & 800\,m & 98.4 [97.6, 99.0] & 98.4 [97.7, 99.1] & 98.3 [97.6, 99.0] & 97.5 [96.7, 98.3] \\
\bottomrule
\end{tabular}

\end{table}

\begin{table}[!htbp]
  \centering
  \footnotesize
  \setlength{\tabcolsep}{3pt}
  \caption{What the null results rule out: paired differences against DINOv2 ViT-B/14 on the unseen-photographer queries (points). An interval's upper end is the largest gain the comparison could be hiding.}
  \label{tab:s-null}
  \begin{tabular}{@{}lcc@{}}
\toprule
Against ViT-B/14 & R@1 & R@10 \\
\midrule
ViT-L/14 & 2.6 [0.5, 4.6] & 0.2 [$-$1.1, 1.6] \\
ViT-B/14 + registers & $-$1.3 [$-$3.2, 0.5] & $-$1.3 [$-$2.6, 0.0] \\
ViT-L/14 + registers & $-$1.0 [$-$3.3, 1.3] & $-$1.8 [$-$3.5, $-$0.2] \\
\bottomrule
\end{tabular}

\end{table}

\clearpage
\section{Photograph credits}
\label{sec:s-credits}

Every photograph reproduced in the main text and in this appendix comes from Wikimedia Commons and remains under the licence its photographer chose; this paper relicenses none of them. All are cropped and resized, and the figures add frames, labels and, in \cref{fig:s-matches}, drawn correspondences. The table lists each photograph once, with the figures that show it (M: a main-text figure, S: an appendix figure), its photographer as credited on Commons, its licence (linked to the licence text) and a link to its Commons page. The licences cover the photographs and not the sculptures depicted.

{\footnotesize
\begin{longtable}{@{}>{\raggedright\arraybackslash}p{2.3cm}>{\raggedright\arraybackslash}p{3.3cm}>{\raggedright\arraybackslash}p{3.9cm}l>{\raggedright\arraybackslash}p{4.6cm}@{}}
\toprule
Shown in & Statue & Photographer & Licence & Commons file \\
\midrule
\endhead
M1, M2, M4, S2, S4 & Słupniki Oławskie & User: (WT-shared) Krasnoludek at wts wikivoyage & \href{https://creativecommons.org/publicdomain/mark/1.0/}{Public domain} & \href{https://commons.wikimedia.org/wiki/File:Krasnal_slupnik_olawska_01.jpg}{Krasnal slupnik olawska 01.jpg} \\
M1, S4 & Słupniki Solne & Pnapora & \href{https://creativecommons.org/licenses/by-sa/3.0}{CC BY-SA 3.0} & \href{https://commons.wikimedia.org/wiki/File:Slupniki_Solne_(Salt_Columnists)_Wroclaw_(1)_dwarf_02.JPG}{Slupniki Solne (Salt Columnist...} \\
M1, S4 & Słupniki Solne & Adalbertus & \href{https://creativecommons.org/licenses/by-sa/4.0}{CC BY-SA 4.0} & \href{https://commons.wikimedia.org/wiki/File:Wroclaw_krasnal_Slupnik_Solny.JPG}{Wroclaw krasnal Slupnik Solny.JPG} \\
M1 & Słupniki Świdnickie & Pnapora & \href{https://creativecommons.org/licenses/by-sa/3.0}{CC BY-SA 3.0} & \href{https://commons.wikimedia.org/wiki/File:Slupniki_Swidnickie_(Swidnicka_Columnists)_Wroclaw_(1)_dwarf_02.JPG}{Slupniki Swidnickie (Swidnicka...} \\
M2 & Bartonik & Fallaner & \href{https://creativecommons.org/licenses/by-sa/4.0}{CC BY-SA 4.0} & \href{https://commons.wikimedia.org/wiki/File:Bartonik_(Barton)_Wroclaw_dwarf_04_2020.jpg}{Bartonik (Barton) Wroclaw dwar...} \\
M2 & Pomagajek & Photographed by Bobby Bothmann. & \href{https://creativecommons.org/licenses/by-sa/4.0}{CC BY-SA 4.0} & \href{https://commons.wikimedia.org/wiki/File:Pomagajek.jpg}{Pomagajek.jpg} \\
M2 & Puszczający Stateczki & Pnapora & \href{https://creativecommons.org/licenses/by-sa/3.0}{CC BY-SA 3.0} & \href{https://commons.wikimedia.org/wiki/File:Krasnale_Wodne_(Water_dwarves)_Puszczaj\%C4\%85cy_Stateczki_(Paper-shi-maker)_Wroclaw_dwarf_02.JPG}{Krasnale Wodne (Water dwarves)...} \\
M2 & Słupniki Solne & User: (WT-shared) Krasnoludek at wts wikivoyage & \href{https://creativecommons.org/publicdomain/mark/1.0/}{Public domain} & \href{https://commons.wikimedia.org/wiki/File:Krasnal_slupnik_solny_01.jpg}{Krasnal slupnik solny 01.jpg} \\
M2 & Zbierający Wodę & Pnapora & \href{https://creativecommons.org/licenses/by-sa/3.0}{CC BY-SA 3.0} & \href{https://commons.wikimedia.org/wiki/File:Krasnale_Wodne_(Water_dwarves)_Zbieraj\%C4\%85cy_Wod\%C4\%99_(Water-collector)_Wroclaw_dwarf_02.JPG}{Krasnale Wodne (Water dwarves)...} \\
M4, S2 & Boliząbek & Pnapora & \href{https://creativecommons.org/licenses/by-sa/4.0}{CC BY-SA 4.0} & \href{https://commons.wikimedia.org/wiki/File:Bolizabek_(Toothache)_Wroclaw_dwarf_04.jpg}{Bolizabek (Toothache) Wroclaw ...} \\
M4, S2 & Drukarz Kacper & Yair Haklai & \href{https://creativecommons.org/licenses/by-sa/4.0}{CC BY-SA 4.0} & \href{https://commons.wikimedia.org/wiki/File:Drukarz_Kacper-Wroc\%C5\%82aw.jpg}{Drukarz Kacper-Wrocław.jpg} \\
M4, S2 & Drukarz Kacper & Fallaner & \href{https://creativecommons.org/licenses/by-sa/4.0}{CC BY-SA 4.0} & \href{https://commons.wikimedia.org/wiki/File:Drukarz_Kacper_(Caspar_the_Printer)_Wroclaw_dwarf_2018_P01.jpg}{Drukarz Kacper (Caspar the Pri...} \\
M4, S2 & Drukarz Kacper & Bernard Ladenthin & \href{https://creativecommons.org/licenses/by/4.0}{CC BY 4.0} & \href{https://commons.wikimedia.org/wiki/File:Drukarz_Kacper_(Caspar_the_Printer)_2019-12-23_15-31-37_Wroc\%C5\%82aw_dwarf.jpg}{Drukarz Kacper (Caspar the Pri...} \\
M4, S2 & Kolanki & Fallaner & \href{https://creativecommons.org/licenses/by-sa/4.0}{CC BY-SA 4.0} & \href{https://commons.wikimedia.org/wiki/File:Kolanki_(Knees)_Wroclaw_dwarf_2020_P03.jpg}{Kolanki (Knees) Wroclaw dwarf ...} \\
M4, S2 & Quadeusz & Fallaner & \href{https://creativecommons.org/licenses/by-sa/4.0}{CC BY-SA 4.0} & \href{https://commons.wikimedia.org/wiki/File:Quadeusz_(Quaddeus)_Wroclaw_dwarf_2020_P03.jpg}{Quadeusz (Quaddeus) Wroclaw dw...} \\
M4, S2 & Słupniki Oławskie & Yair Haklai & \href{https://creativecommons.org/licenses/by-sa/4.0}{CC BY-SA 4.0} & \href{https://commons.wikimedia.org/wiki/File:Dwarf-Wroc\%C5\%82aw-3.jpg}{Dwarf-Wrocław-3.jpg} \\
M4, S2 & Zbierający Wodę & Pnapora & \href{https://creativecommons.org/licenses/by-sa/3.0}{CC BY-SA 3.0} & \href{https://commons.wikimedia.org/wiki/File:Krasnale_Wodne_(Water_dwarves)_Zbieraj\%C4\%85cy_Wod\%C4\%99_(Water-collector)_Wroclaw_dwarf_01.JPG}{Krasnale Wodne (Water dwarves)...} \\
S2 & Antykwarek & Pnapora & \href{https://creativecommons.org/licenses/by-sa/3.0}{CC BY-SA 3.0} & \href{https://commons.wikimedia.org/wiki/File:Antykwarek_(Anitquarian)_Wroclaw_dwarf_01.JPG}{Antykwarek (Anitquarian) Wrocl...} \\
S2 & Bagażowy & Agnieszka Kwiecień, Nova & \href{https://creativecommons.org/licenses/by-sa/4.0}{CC BY-SA 4.0} & \href{https://commons.wikimedia.org/wiki/File:Baga\%C5\%BCowy_2019-06-07_Wroc\%C5\%82aw_02.jpg}{Bagażowy 2019-06-07 Wrocław 02...} \\
S2 & Bibliofil & Pnapora & \href{https://creativecommons.org/licenses/by-sa/3.0}{CC BY-SA 3.0} & \href{https://commons.wikimedia.org/wiki/File:Bibliofil_(Booklover)_Wroclaw_dwarf_01.JPG}{Bibliofil (Booklover) Wroclaw ...} \\
S2 & Capgeminiusz Programista & Fallaner (crop by Manfred\_Werner\_(WMAT) ) & \href{https://creativecommons.org/licenses/by-sa/4.0}{CC BY-SA 4.0} & \href{https://commons.wikimedia.org/wiki/File:Korpokrasnal_(Company_dwarf)_Wroclaw_dwarf_02_crop.jpg}{Korpokrasnal (Company dwarf) W...} \\
S2 & Concierge dwarf at the entrance of Wroclavia Shopping Center (Grzegorz Łagowski) & Fallaner & \href{https://creativecommons.org/licenses/by-sa/4.0}{CC BY-SA 4.0} & \href{https://commons.wikimedia.org/wiki/File:Concierge_(Concierge)_Wroclaw_dwarf_04.jpg}{Concierge (Concierge) Wroclaw ...} \\
S2 & Dialogomir & MCDBMV & \href{https://creativecommons.org/licenses/by-sa/3.0}{CC BY-SA 3.0} & \href{https://commons.wikimedia.org/wiki/File:\%E3\%83\%B4\%E3\%83\%AD\%E3\%83\%84\%E3\%83\%AF\%E3\%83\%95\%E3\%81\%AE\%E5\%B0\%8F\%E4\%BA\%BA\%EF\%BC\%88\%E3\%82\%BD\%E3\%83\%AB\%E3\%83\%88\%E5\%BA\%83\%E5\%A0\%B4\%EF\%BC\%89.JPG}{Commons file page} \\
S2 & Eldżiki & Pnapora & \href{https://creativecommons.org/licenses/by-sa/3.0}{CC BY-SA 3.0} & \href{https://commons.wikimedia.org/wiki/File:Eldziki_Tabus_i_Assik_Wroclaw_dwarf_02.JPG}{Eldziki Tabus i Assik Wroclaw ...} \\
S2 & Fatuś & Pnapora & \href{https://creativecommons.org/licenses/by-sa/3.0}{CC BY-SA 3.0} & \href{https://commons.wikimedia.org/wiki/File:Fatus_(Lathe_operator)_Wroclaw_dwarf_02.JPG}{Fatus (Lathe operator) Wroclaw...} \\
S2 & Gaviusz & Pnapora & \href{https://creativecommons.org/licenses/by-sa/4.0}{CC BY-SA 4.0} & \href{https://commons.wikimedia.org/wiki/File:Gaviusz_(Gavius)_Wroclaw_dwarf_03.JPG}{Gaviusz (Gavius) Wroclaw dwarf...} \\
S2 & Golasek dwarf in Wrocław & Fallaner & \href{https://creativecommons.org/licenses/by-sa/4.0}{CC BY-SA 4.0} & \href{https://commons.wikimedia.org/wiki/File:Golasek_(Naked_Dwarf)_Wroclaw_dwarf_2020_P06.jpg}{Golasek (Naked Dwarf) Wroclaw ...} \\
S2 & Implantolog & Fallaner & \href{https://creativecommons.org/licenses/by-sa/4.0}{CC BY-SA 4.0} & \href{https://commons.wikimedia.org/wiki/File:Krasnal_z_Limedu_(Dwarf_from_Limed)_Wroclaw_dwarf_02.jpg}{Krasnal z Limedu (Dwarf from L...} \\
S2 & Kapitan Jack Port & Pnapora & \href{https://creativecommons.org/licenses/by-sa/4.0}{CC BY-SA 4.0} & \href{https://commons.wikimedia.org/wiki/File:Pirat_(Pirate)_Wroclaw_dwarf_03.jpg}{Pirat (Pirate) Wroclaw dwarf 0...} \\
S2 & Karmiący Ptaki & Pnapora & \href{https://creativecommons.org/licenses/by-sa/3.0}{CC BY-SA 3.0} & \href{https://commons.wikimedia.org/wiki/File:Krasnale_Wodne_(Water_dwarves)_Karmi\%C4\%85cy_Ptaki_(Bird-feeder)_Wroclaw_dwarf_02.JPG}{Krasnale Wodne (Water dwarves)...} \\
S2 & Koncertuś & Fallaner & \href{https://creativecommons.org/licenses/by-sa/4.0}{CC BY-SA 4.0} & \href{https://commons.wikimedia.org/wiki/File:Koncertus_(Cencerter)_Wroclaw_dwarf_2020_P02.jpg}{Koncertus (Cencerter) Wroclaw ...} \\
S2 & Ortopeda & Pnapora & \href{https://creativecommons.org/licenses/by-sa/4.0}{CC BY-SA 4.0} & \href{https://commons.wikimedia.org/wiki/File:Ortopeda_(Orthopaedist)_Wroclaw_dwarf_03.jpg}{Ortopeda (Orthopaedist) Wrocla...} \\
S2 & Pomnik Pomarańczowej Alternatywy & Pnapora & \href{https://creativecommons.org/licenses/by-sa/3.0}{CC BY-SA 3.0} & \href{https://commons.wikimedia.org/wiki/File:Papa_Krasnal_(Papa_Dwarf)_Wroclaw_dwarf_02.JPG}{Papa Krasnal (Papa Dwarf) Wroc...} \\
S2 & RMFek & Pnapora & \href{https://creativecommons.org/licenses/by-sa/4.0}{CC BY-SA 4.0} & \href{https://commons.wikimedia.org/wiki/File:Sluchacz_RMF_(RMF_Listener)_Wroclaw_dwarf_03.jpg}{Sluchacz RMF (RMF Listener) Wr...} \\
S2 & RMFek & Pnapora & \href{https://creativecommons.org/licenses/by-sa/4.0}{CC BY-SA 4.0} & \href{https://commons.wikimedia.org/wiki/File:Sluchacz_RMF_(RMF_Listener)_Wroclaw_dwarf_01.jpg}{Sluchacz RMF (RMF Listener) Wr...} \\
S2 & Siłacz & Fallaner & \href{https://creativecommons.org/licenses/by-sa/4.0}{CC BY-SA 4.0} & \href{https://commons.wikimedia.org/wiki/File:Silacz_(Strongmen)_Wroclaw_dwarf_2020_P03.jpg}{Silacz (Strongmen) Wroclaw dwa...} \\
S2 & Słupniki Oławskie & Pnapora & \href{https://creativecommons.org/licenses/by-sa/3.0}{CC BY-SA 3.0} & \href{https://commons.wikimedia.org/wiki/File:Slupniki_Olawskie_(Olawska_Columnists)_Wroclaw_(3)_dwarf_02.jpg}{Slupniki Olawskie (Olawska Col...} \\
S2 & Słupniki Solne & Pnapora & \href{https://creativecommons.org/licenses/by-sa/3.0}{CC BY-SA 3.0} & \href{https://commons.wikimedia.org/wiki/File:Slupniki_Solne_(Salt_Columnists)_Wroclaw_(4)_dwarf_01.JPG}{Slupniki Solne (Salt Columnist...} \\
S2 & Słupniki Świdnickie & Pnapora & \href{https://creativecommons.org/licenses/by-sa/3.0}{CC BY-SA 3.0} & \href{https://commons.wikimedia.org/wiki/File:Slupniki_Swidnickie_(Swidnicka_Columnists)_Wroclaw_(1)_dwarf_01.JPG}{Slupniki Swidnickie (Swidnicka...} \\
S2 & Turysta & Pnapora & \href{https://creativecommons.org/licenses/by-sa/3.0}{CC BY-SA 3.0} & \href{https://commons.wikimedia.org/wiki/File:Turysta_(Tourist)_Wroclaw_dwarf_01.JPG}{Turysta (Tourist) Wroclaw dwar...} \\
S2 & WIZ-ek & Fallaner & \href{https://creativecommons.org/licenses/by-sa/4.0}{CC BY-SA 4.0} & \href{https://commons.wikimedia.org/wiki/File:WIZek_(WIZ)_Wroclaw_dwarf_2021_P01.jpg}{WIZek (WIZ) Wroclaw dwarf 2021...} \\
S4 & Pacjent i Lekarz & Fallaner & \href{https://creativecommons.org/licenses/by-sa/4.0}{CC BY-SA 4.0} & \href{https://commons.wikimedia.org/wiki/File:Pacjent_i_lekarz_(Patient_and_Doc)_Wroclaw_dwarf_2020_P05.jpg}{Pacjent i lekarz (Patient and ...} \\
S4 & Pacjent i Lekarz & Pnapora & \href{https://creativecommons.org/licenses/by-sa/3.0}{CC BY-SA 3.0} & \href{https://commons.wikimedia.org/wiki/File:Lekarz_i_pacjent_(Patient_and_Doc)_Wroclaw_dwarf_01.JPG}{Lekarz i pacjent (Patient and ...} \\
S4 & Telesfor & Pnapora & \href{https://creativecommons.org/licenses/by-sa/3.0}{CC BY-SA 3.0} & \href{https://commons.wikimedia.org/wiki/File:Telesfor_(Telesfor)_Wroclaw_dwarf_01.JPG}{Telesfor (Telesfor) Wroclaw dw...} \\
\bottomrule
\end{longtable}

}

\end{document}